\documentclass[11pt]{article}

\usepackage[final]{acl}

\usepackage{times}
\usepackage{latexsym}
\usepackage[T1]{fontenc}
\usepackage[utf8]{inputenc}
\usepackage{microtype}
\usepackage{inconsolata}
\usepackage{graphicx}
\usepackage{placeins}
\usepackage{etoc}

\usepackage{booktabs}
\usepackage{longtable}   
\usepackage{amsmath}
\usepackage{amssymb}
\usepackage{multirow}
\usepackage{enumitem}
\usepackage{xcolor}
\usepackage{algorithm}
\usepackage{algpseudocode}
\usepackage{tikz}
\usetikzlibrary{arrows.meta,positioning,fit,backgrounds,calc}

\newcommand{\heb}{\textsc{HarnessOpt-Bench}\xspace}
\usepackage{xspace}
\newcommand{\vero}{\textsc{VeRO}\xspace}

\definecolor{pos}{RGB}{0,120,0}
\definecolor{neg}{RGB}{170,0,0}

\newcommand{\cc}{\texttt{claude-code}\xspace}
\newcommand{\oc}{\texttt{opencode}\xspace}
\newcommand{\kc}{\texttt{kimi-cli}\xspace}
\newcommand{\codex}{\texttt{codex}\xspace}

\newcommand{\miniswe}{\texttt{mini-swe-agent}\xspace}

\title{HarnessOpt-Bench: Evaluating LLMs at Harness Optimization}

\author{
  \textbf{Varun Ursekar},
  \textbf{Apaar Shanker},
  \textbf{Yash Maurya},
  \textbf{Shehab Yasser},
  \textbf{Vijay S Kalmath},
\\
  \textbf{Veronica Chatrath},
  \textbf{Yuan Xue}
\\
  Scale AI
\\
  \small{\textbf{Correspondence:} \href{mailto:varun.ursekar@scale.com}{varun.ursekar@scale.com}}
}

\definecolor{varun}{RGB}{100,100,0}

\newcommand{\NumPrimaryRuns}{180\xspace}
\newcommand{\NumPrimaryTaskCells}{60\xspace}
\newcommand{\NumPrimaryConfigurations}{15\xspace}

\newcommand{\NumPaperTasks}{4\xspace}
\newcommand{\NumSuiteTasks}{4\xspace}
\newcommand{\NumRounds}{3\xspace}
\newcommand{\NumModelPairs}{10\xspace}
\newcommand{\NumResolvedModelPairs}{7\xspace}
\newcommand{\NumRankResamples}{50000\xspace}
\newcommand{\NumNativeOpenCodePairs}{20\xspace}
\newcommand{\NativeWinsOpenCode}{10\xspace}
\newcommand{\OpenCodeWins}{10\xspace}

\newcommand{\NumNativeMiniSwePairs}{20\xspace}
\newcommand{\NativeWinsMiniSwe}{6\xspace}
\newcommand{\MiniSweWins}{13\xspace}

\newcommand{\NumTransferPrimaryCandidates}{180\xspace}

\newcommand{\NumTransferResolvedPairs}{8\xspace}
\newcommand{\NumTransferRetainedPairs}{7\xspace}

\newcommand{\NumSeedGridRuns}{270\xspace}

\newcommand{\ComputeGainPerOutputDoubling}{-0.005\xspace}
\newcommand{\ComputeGainLo}{-0.055\xspace}
\newcommand{\ComputeGainHi}{+0.029\xspace}

\newcommand{\NumLadderSeries}{6\xspace}

\newcommand{\NumLadderAdditionalRuns}{72\xspace}

\newcommand{\LadderRhoLo}{0.90\xspace}
\newcommand{\LadderRhoHi}{1.00\xspace}

\newcommand{\NumProcessRuns}{180\xspace}

\newcommand{\LeverRhoLo}{+0.40\xspace}
\newcommand{\LeverRhoHi}{+0.80\xspace}
\newcommand{\ChurnRhoLo}{+0.50\xspace}
\newcommand{\ChurnRhoHi}{+0.88\xspace}
\newcommand{\TraceRhoLo}{-0.75\xspace}
\newcommand{\TraceRhoHi}{-0.42\xspace}
\newcommand{\NSpanCalls}{21\xspace}
\newcommand{\NSpanCells}{7\xspace}
\newcommand{\NumValidationObserved}{162\xspace}

\newcommand{\NumTestBelowValidation}{115\xspace}
\newcommand{\NBudgetRuns}{180\xspace}
\newcommand{\EvalCallsUsed}{8\xspace}
\newcommand{\EvalCallCapPerPartition}{100\xspace}
\newcommand{\EvalCallCapTotal}{200\xspace}
\newcommand{\EvalCallFrac}{4\%\xspace}
\newcommand{\CasePassFrac}{83\%\xspace}
\newcommand{\NBudgetExhausted}{101\xspace}

\DeclareRobustCommand{\add}[1]{#1}

\let\revNumPrimaryRuns\NumPrimaryRuns\renewcommand{\NumPrimaryRuns}{\add{\revNumPrimaryRuns}\xspace}
\let\revNumPrimaryTaskCells\NumPrimaryTaskCells\renewcommand{\NumPrimaryTaskCells}{\add{\revNumPrimaryTaskCells}\xspace}
\let\revNumPrimaryConfigurations\NumPrimaryConfigurations\renewcommand{\NumPrimaryConfigurations}{\add{\revNumPrimaryConfigurations}\xspace}

\let\revNumRounds\NumRounds\renewcommand{\NumRounds}{\add{\revNumRounds}\xspace}
\let\revNumModelPairs\NumModelPairs\renewcommand{\NumModelPairs}{\add{\revNumModelPairs}\xspace}
\let\revNumResolvedModelPairs\NumResolvedModelPairs\renewcommand{\NumResolvedModelPairs}{\add{\revNumResolvedModelPairs}\xspace}
\let\revNumRankResamples\NumRankResamples\renewcommand{\NumRankResamples}{\add{\revNumRankResamples}\xspace}
\let\revNumNativeOpenCodePairs\NumNativeOpenCodePairs\renewcommand{\NumNativeOpenCodePairs}{\add{\revNumNativeOpenCodePairs}\xspace}
\let\revNativeWinsOpenCode\NativeWinsOpenCode\renewcommand{\NativeWinsOpenCode}{\add{\revNativeWinsOpenCode}\xspace}
\let\revOpenCodeWins\OpenCodeWins\renewcommand{\OpenCodeWins}{\add{\revOpenCodeWins}\xspace}

\let\revNumNativeMiniSwePairs\NumNativeMiniSwePairs\renewcommand{\NumNativeMiniSwePairs}{\add{\revNumNativeMiniSwePairs}\xspace}
\let\revNativeWinsMiniSwe\NativeWinsMiniSwe\renewcommand{\NativeWinsMiniSwe}{\add{\revNativeWinsMiniSwe}\xspace}
\let\revMiniSweWins\MiniSweWins\renewcommand{\MiniSweWins}{\add{\revMiniSweWins}\xspace}

\let\revNumTransferPrimaryCandidates\NumTransferPrimaryCandidates\renewcommand{\NumTransferPrimaryCandidates}{\add{\revNumTransferPrimaryCandidates}\xspace}

\let\revNumTransferResolvedPairs\NumTransferResolvedPairs\renewcommand{\NumTransferResolvedPairs}{\add{\revNumTransferResolvedPairs}\xspace}
\let\revNumTransferRetainedPairs\NumTransferRetainedPairs\renewcommand{\NumTransferRetainedPairs}{\add{\revNumTransferRetainedPairs}\xspace}

\let\revNumSeedGridRuns\NumSeedGridRuns\renewcommand{\NumSeedGridRuns}{\add{\revNumSeedGridRuns}\xspace}

\let\revComputeGainPerOutputDoubling\ComputeGainPerOutputDoubling\renewcommand{\ComputeGainPerOutputDoubling}{\add{\revComputeGainPerOutputDoubling}\xspace}
\let\revComputeGainLo\ComputeGainLo\renewcommand{\ComputeGainLo}{\add{\revComputeGainLo}\xspace}
\let\revComputeGainHi\ComputeGainHi\renewcommand{\ComputeGainHi}{\add{\revComputeGainHi}\xspace}

\let\revNumLadderSeries\NumLadderSeries\renewcommand{\NumLadderSeries}{\add{\revNumLadderSeries}\xspace}

\let\revNumLadderAdditionalRuns\NumLadderAdditionalRuns\renewcommand{\NumLadderAdditionalRuns}{\add{\revNumLadderAdditionalRuns}\xspace}

\let\revLadderRhoLo\LadderRhoLo\renewcommand{\LadderRhoLo}{\add{\revLadderRhoLo}\xspace}
\let\revLadderRhoHi\LadderRhoHi\renewcommand{\LadderRhoHi}{\add{\revLadderRhoHi}\xspace}

\let\revNumProcessRuns\NumProcessRuns\renewcommand{\NumProcessRuns}{\add{\revNumProcessRuns}\xspace}

\let\revLeverRhoLo\LeverRhoLo\renewcommand{\LeverRhoLo}{\add{\revLeverRhoLo}\xspace}
\let\revLeverRhoHi\LeverRhoHi\renewcommand{\LeverRhoHi}{\add{\revLeverRhoHi}\xspace}
\let\revChurnRhoLo\ChurnRhoLo\renewcommand{\ChurnRhoLo}{\add{\revChurnRhoLo}\xspace}
\let\revChurnRhoHi\ChurnRhoHi\renewcommand{\ChurnRhoHi}{\add{\revChurnRhoHi}\xspace}
\let\revTraceRhoLo\TraceRhoLo\renewcommand{\TraceRhoLo}{\add{\revTraceRhoLo}\xspace}
\let\revTraceRhoHi\TraceRhoHi\renewcommand{\TraceRhoHi}{\add{\revTraceRhoHi}\xspace}
\let\revNSpanCalls\NSpanCalls\renewcommand{\NSpanCalls}{\add{\revNSpanCalls}\xspace}
\let\revNumValidationObserved\NumValidationObserved\renewcommand{\NumValidationObserved}{\add{\revNumValidationObserved}\xspace}

\let\revNumTestBelowValidation\NumTestBelowValidation\renewcommand{\NumTestBelowValidation}{\add{\revNumTestBelowValidation}\xspace}
\let\revNBudgetRuns\NBudgetRuns\renewcommand{\NBudgetRuns}{\add{\revNBudgetRuns}\xspace}
\let\revEvalCallCapPerPartition\EvalCallCapPerPartition\renewcommand{\EvalCallCapPerPartition}{\add{\revEvalCallCapPerPartition}\xspace}
\let\revEvalCallCapTotal\EvalCallCapTotal\renewcommand{\EvalCallCapTotal}{\add{\revEvalCallCapTotal}\xspace}
\let\revCasePassFrac\CasePassFrac\renewcommand{\CasePassFrac}{\add{\revCasePassFrac}\xspace}
\let\revNBudgetExhausted\NBudgetExhausted\renewcommand{\NBudgetExhausted}{\add{\revNBudgetExhausted}\xspace}

\begin{document}
\maketitle
\etocdepthtag.toc{main}

\begin{abstract}
As LLMs are increasingly deployed within \emph{\add{agentic} systems}, their capabilities depend not only on \add{the model} weights but also on \add{the} \emph{harness}: the prompts, tools, control flow, memory, and orchestration code surrounding them. 
\add{This makes automated harness optimization, the improvement of a harness by an AI system,
both a route to better AI systems and a demanding capability in its own right.} We introduce
\heb, a benchmark \add{for} end-to-end harness \add{optimization under expensive and stochastic
evaluation. \heb provides a sandboxed environment in which an optimizer, an LLM in a coding
harness, iteratively edits and evaluates a target agent's harness within a fixed budget, then
submits a final candidate that is scored by normalized gain over the baseline on a held-out
test partition. We} evaluate \add{5 frontier LLMs as optimizers under two shared and their
native coding harnesses} across \NumPaperTasks\ downstream \add{tasks, with
\NumRounds\ runs per condition. Of the \NumModelPairs\ model pairs,
\NumResolvedModelPairs\ differ clearly, and these differences hold when the final candidates are run with two unseen target models.
Native coding harnesses give no consistent advantage. Gains depend on the task and on the
starting harness's strength. These results establish harness optimization as
measurable and discriminative, with space for improvement. Code is available at
\url{https://github.com/scaleapi/vero/tree/main}.}

\end{abstract}
\section{Introduction}

Language models are increasingly deployed within \emph{harnesses}: the programs that specify their prompts, tools, control flow, context and memory, and the orchestration code invoking them
\citep{weng2023agent,wang2024survey,weng2026harness}. The same model can exhibit
substantially different capabilities under different harnesses
\citep{yao2026harnessbenchmeasuringharnesseffects, zhang2026bindingconstraint}, which makes \emph{harness
optimization} -- the iterative improvement of a harness on a
measured outcome under a fixed budget -- an increasingly important part of building capable AI systems. 

Recent work asks whether AI systems can automate this process, extending a
broader line on self-improving agents, agent meta-optimization, and automated
research
\citep{rank2026posttrainbenchllmagentsautomate,
meng2026rsibenchdatabenchmarkingdatacentricresearch,
wijk2025rebenchevaluatingfrontierai}. These approaches \add{often use LLMs, but} differ in the \add{\emph{optimizer harness}: the program that scaffolds the model and drives optimization.} ShinkaEvolve
\citep{lange2025shinkaevolveopenendedsampleefficientprogram} and GEPA
\citep{agrawal2026gepa} \add{sample context and prompt} \add{models to propose variants inside an} evolutionary
\add{search procedure. Meta-Harness
\citep{lee2026metaharnessendtoendoptimizationmodel} instead runs a fixed loop that, at each iteration, has a
multi-turn coding agent propose candidates based on the harness' codebase and traces of earlier
evaluations, then evaluates them.} \vero\ \citep{ursekar2026vero}
\add{hands the end-to-end search to the agent, allowing it to decide what to evaluate and propagate.}

\begin{figure*}[t]
\centering
\includegraphics[width=\linewidth]{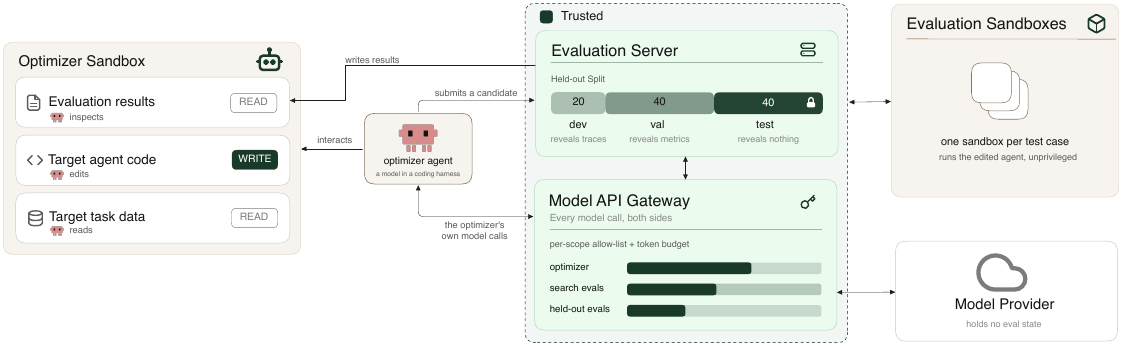}
\caption{\textbf{\add{The \heb\ optimization environment.}} The optimizer can write only the target agent's harness. It can read the evaluation results and the target task data but not modify them. The optimizer
receives per-case development traces and aggregate validation metrics. Each candidate is evaluated in isolated, unprivileged sandboxes, and every model call, both the optimizer's own and the evaluations', passes through a gateway that enforces model allow-lists and \add{meters usage.} The test partition is evaluated only after the optimizer nominates a final candidate. \add{Held-out data, provider credentials, and budget enforcement are absent from the optimizer's sandbox, so} these restrictions are properties of the \add{environment} rather than instructions the optimizer is expected to follow.}
\label{fig:arch}
\end{figure*}

\add{The cost of evaluation shapes the choice of optimizer harness.}
Where evaluation is cheap, \add{trying many candidates} substitutes
for reasoning about any one of them; where it is expensive and noisy, reasoning
from prior evidence becomes \add{central. Estimating the effect of a harness change
entails running a stochastic agent over many cases. This becomes expensive as target agents
and tasks grow in complexity. In that regime optimization must be sample-efficient,
choosing each change from what earlier evaluations revealed about why the harness fails.} Our
benchmark targets \add{such tasks} (fixed-corpus
research, terminal use) rather than ones that are cheap to score.

\add{End-to-end harness optimization is also a capability worth measuring in frontier
models in its own right. It is long-horizon and open-ended (with no single correct solution), and it tests several core research skills, including writing and debugging code,
designing experiments, interpreting stochastic results, and deciding which changes to keep under uncertainty. While benchmarks exist for each of these separately, \heb\ measures them
together, as one capability: the ability to optimize harnesses.}

\add{Every harness optimization method must define an evaluation interface: the functions an optimizer
harness may invoke to evaluate a candidate (on which split, over which cases, how often) and what
each function returns. \heb\ fixes this interface, so optimizers get equal resources and information
and never see held-out evaluations, while leaving the optimizer harness open. We use a generic
coding agent. Unlike optimizer harnesses built specifically for harness optimization, it encodes no search strategy of
its own, leaving the design of the search to the underlying model.}

\add{In this setting we ask
three questions: can frontier models be told apart as harness optimizers} (\textbf{RQ1}),
\add{where do} they fall short (\textbf{RQ2}), and how \add{do results change with} the
optimizer's own harness (\textbf{RQ3})\add{? Alongside RQ1, we check whether
optimized harnesses generalize to unseen target models, how gains change with the strength of
the starting harness, and whether results track successive releases of a model family.}

\add{We make the following contributions:}

\begin{itemize}[leftmargin=*,itemsep=2pt,topsep=3pt]

    \item \add{\textbf{A benchmark environment for harness optimization.} \heb\ provides
    \NumSuiteTasks\ downstream tasks with pinned starting harnesses, fixed development,
    validation, and test splits, and per-split limits on evaluation feedback. The
    environment, built on \vero\ \citep{ursekar2026vero}, is extensible to new tasks. It enforces
    the evaluation budget, keeps the test partition hidden, and versions every candidate (Figure~}\ref{fig:arch}\add{).}

    \item \add{\textbf{A controlled comparison of optimizers.} We evaluate five frontier
    models under two shared coding harnesses and their native harnesses on
    \NumPaperTasks\ tasks, with \NumRounds\ runs per condition. Of the \NumModelPairs\ model pairs, \NumResolvedModelPairs\ differ clearly. These
    differences hold even when optimized harnesses are tested on two unseen target models -- though
    the harnesses keep only part of their gain. Newer model releases generally optimize better
    (Section~}\ref{sec:ladder}\add{). Native coding harnesses give no consistent advantage
    (\textbf{RQ1}, \textbf{RQ3}).}

    \item \add{\textbf{Evidence on what limits current optimizers.} Optimizers improve weak
    starting harnesses far more than strong ones (Section~}\ref{sec:seed-maturity}\add{). Broader
    and larger edits go with higher gain, whereas detailed trace inspection does not (\textbf{RQ2}, Section~}\ref{sec:process}\add{).}

\end{itemize}

\section{Related Work}
\label{sec:related}

\begin{algorithm}[t]
\caption{The \heb\ protocol}
\label{alg:protocol}
\begin{algorithmic}[1]
\Require $H_0$; $\theta=(\mathcal{M},E,V)$; $B$; $\pi$
\State $\mathcal{C}\gets\{H_0\}$
\While{$\sum_j c_j\leq B$ and no candidate is nominated}
  \State \textbf{either}: commit $H'\in\mathcal{H}$ and add it to $\mathcal{C}$
  \State \textbf{or}: choose $H\in\mathcal{C}$ and cases $Q$
  \State $(\hat{s},\varphi)\gets F_{\theta}(H,Q)$
  \State observe $\pi_{\mathcal{D}}(\hat{s},\varphi)$
\EndWhile
\State nominate $H^{+}\in\mathcal{C}$
\State server evaluates $H^{+}$ on $\mathcal{D}^{\mathrm{test}}$
\State \textbf{report} $g$ from Eq.~\ref{eq:normalized-gain}
\end{algorithmic}
\end{algorithm}

\paragraph{Automated code and harness optimization.}

A long line of work, encompassing OPRO, FunSearch, AlphaEvolve, and ShinkaEvolve, has applied LLMs within larger search \add{procedures} to discover novel or optimized programs 
\citep{yang2024opro,romera2024funsearch,novikov2025alphaevolve,lange2025shinkaevolveopenendedsampleefficientprogram}. 

Existing approaches to \emph{harness optimization} differ along two axes: which parts of the harness they may
modify, and the role assigned to the LLM during search. Prompt optimization
methods such as DSPy \citep{khattab2024dspy} optimize prompts while holding the surrounding program fixed. Meta Context Engineering
\citep{ye2026metacontextengineeringagentic} searches over \emph{skills} and context
artifacts in a bi-level procedure. TextGrad
\citep{yuksekgonul2025textgrad}, Trace  \citep{cheng2024trace}, and LLM-AutoDiff
\citep{yin2025adalflow} optimize arbitrary text components in chained workflows by
back-propagating textual feedback using LLMs. GEPA \citep{agrawal2026gepa} optimizes textual artifacts by using
reflective feedback to guide an evolutionary process. 

A second line expands the search space to the
agent program itself: STOP \citep{zelikman2024stop} recursively improves a
scaffolding program. ADAS \citep{hu2025adas} searches over agent designs
represented as code. AFlow \citep{zhang2025aflow} performs tree search over
code-defined workflows. Darwin
G\"odel Machine \citep{zhang2026dgm}, G\"odel Agent
\citep{yin2025godelagent}, and SICA
\citep{robeyns2025sica} study recursive \emph{self-modification}.

Closest to the regime \heb\ measures are optimizers \add{built around a coding agent}:
\citet{ursekar2026vero} and \citet{lee2026metaharnessendtoendoptimizationmodel} let a coding agent read the target source \add{and the}
scores and execution traces \add{of} prior candidates, and \add{edit the harness as a codebase
rather than proposing mutations for a fixed search algorithm.} 

Other
systems in this regime include Agentic Harness Engineering
\citep{lin2026agenticharness} and HarnessX
\citep{chen2026harnessx}. Several of these works contribute optimization
\emph{methods}, evaluated on method-specific \add{starting harnesses}, budgets, search
spaces, and scoring protocols, which makes their reported outcomes difficult to
compare. \heb\ is complementary: it fixes the optimization problem and the
evaluation protocol so that optimizers can
be compared on common ground. \heb\ \add{builds on \vero's} infrastructure and protocol \add{with} a suite of optimization tasks. \add{Our experiments adopt \vero's setting, in which the optimizer controls the whole loop.} 

\paragraph{Coding agent benchmarks.}
Coding agent evaluation has grown from function-level synthesis
\citep{chen2021humaneval,austin2021mbpp} to repository-scale tasks in real
execution environments \citep{jimenez2024swebench}, and tasks requiring iterative optimization, such as machine learning
engineering \citep{chan2025mlebench} and kernel optimization \citep{ouyang2025kernelbench}. \heb\ similarly requires coding agents to navigate the target system codebase and iterate on environmental feedback; it differs in what the optimization target is (i.e. a harness), the noisiness of evaluation feedback, and the types of bounds imposed on its search.

\section{\heb: Harness Optimization as a Task}
\label{sec:task}

\add{
\begin{table*}[t]
\centering\footnotesize
\setlength{\tabcolsep}{4pt}
\resizebox{\textwidth}{!}{%
\begin{tabular}{@{}lccccccc@{}}
\toprule
 & & & \multicolumn{5}{c}{\textbf{Off-the-shelf harness}}\\
\cmidrule(l){4-8}
\textbf{Task} & \textbf{Split (d/v/t)} & \textbf{Starting} & \textbf{\texttt{opencode}} & \textbf{\texttt{goose}} & \textbf{\texttt{openhands-sdk}} & \textbf{\texttt{mini-swe-agent}} & \textbf{\texttt{terminus-2}}\\
\midrule
\shortstack[l]{OfficeQA\\[1pt]{\tiny \texttt{deepseek-v4-flash}}} & $49/98/99$ & \shortstack[c]{$0.341$\\[1pt]{\tiny $\pm0.023$}} & \shortstack[c]{$0.727$\\[1pt]{\tiny $\pm0.010$}} & \shortstack[c]{$0.505$\\[1pt]{\tiny $\pm0.129$}} & \shortstack[c]{$0.713$\\[1pt]{\tiny $\pm0.011$}} & \shortstack[c]{$\mathbf{0.734}$\\[1pt]{\tiny $\pm0.009$}} & \shortstack[c]{$0.687$\\[1pt]{\tiny $\pm0.031$}}\\
\addlinespace[3pt]
\shortstack[l]{BrowseComp-Plus\\[1pt]{\tiny \texttt{deepseek-v4-flash}}} & $33/66/66$ & \shortstack[c]{$0.462$\\[1pt]{\tiny $\pm0.020$}} & \shortstack[c]{$0.434$\\[1pt]{\tiny $\pm0.018$}} & \shortstack[c]{$0.615$\\[1pt]{\tiny $\pm0.013$}} & \shortstack[c]{$0.657$\\[1pt]{\tiny $\pm0.035$}} & \shortstack[c]{$\mathbf{0.701}$\\[1pt]{\tiny $\pm0.009$}} & \shortstack[c]{$0.439$\\[1pt]{\tiny $\pm0.023$}}\\
\addlinespace[3pt]
\shortstack[l]{Terminal-Bench\\[1pt]{\tiny \texttt{grok-build}}} & $17/36/36$ & \shortstack[c]{$0.241$\\[1pt]{\tiny $\pm0.009$}} & \shortstack[c]{$\mathbf{0.607}$\\[1pt]{\tiny $\pm0.030$}} & \shortstack[c]{$0.434$\\[1pt]{\tiny $\pm0.024$}} & \shortstack[c]{$0.393$\\[1pt]{\tiny $\pm0.013$}} & \shortstack[c]{$0.364$\\[1pt]{\tiny $\pm0.032$}} & \shortstack[c]{$0.333$\\[1pt]{\tiny $\pm0.016$}}\\
\addlinespace[3pt]
\shortstack[l]{GAIA\\[1pt]{\tiny \texttt{gpt-5.4-mini}}} & $33/66/66$ & $0.000$ & \shortstack[c]{$0.469$\\[1pt]{\tiny $\pm0.023$}} & \shortstack[c]{$0.207$\\[1pt]{\tiny $\pm0.013$}} & \shortstack[c]{$\mathbf{0.508}$\\[1pt]{\tiny $\pm0.029$}} & \shortstack[c]{$0.172$\\[1pt]{\tiny $\pm0.022$}} & \shortstack[c]{$0.202$\\[1pt]{\tiny $\pm0.005$}}\\
\bottomrule
\end{tabular}}
\caption{Suite tasks, target models (second line), development/validation/test splits, starting harness scores, and off-the-shelf harness scores. Scores are mean $\pm$ one standard error over $K{=}3$ evaluation rounds. The best off-the-shelf harness per task is bold. GAIA's starting harness is a non-functional stub that is designed to score zero. Timeouts score zero. The off-the-shelf harnesses give context for the starting harness score.}
\label{tab:suite}
\end{table*}
}

Harness optimization is a constrained, stochastic
program optimization problem. Algorithm~\ref{alg:protocol} summarizes the
interaction protocol an optimizer must satisfy; this section defines each element in its
general form, then fixes it for this work.

\subsection{The optimization problem}

\paragraph{Candidates and invariants.}
A candidate harness $H$ is an executable codebase. No semantic partition is imposed between
prompts, tool definitions, memory, and control flow. The optimizer may edit, add,
or delete files subject to a fixed execution interface and a small set of
immutable paths. We write $\mathcal{H}$ for the feasible set. A \add{\emph{task} fixes a seed harness
$H_0\in\mathcal{H}$, from which optimization starts, and} invariants $\theta=(\mathcal{M},E,V)$:
the models $\mathcal{M}$ available to a candidate, the environment $E(x)$ associated with each case $x$, and the verifier $V$ mapping a completed trajectory to a score in $[0,1]$. The optimizer may change $H$ but \add{neither $H_0$ nor} $\theta$; changing \add{either} defines a different task, not a different candidate. A harness' web access, for example, might be controlled by $E$. 

\paragraph{Evaluation and disclosure.}
Cases are partitioned into disjoint development, validation, and test sets,
$\mathcal{D}^{\mathrm{dev}}$, $\mathcal{D}^{\mathrm{val}}$, and
$\mathcal{D}^{\mathrm{test}}$. Executing harness $H$ on case $x$ produces a
stochastic trajectory
$\tau\sim\operatorname{Rollout}(H,\theta,x)$, to which the verifier assigns
score $V(\tau,x)$. We define the expected score on partition $\mathcal{D}$ as
\begin{equation}
\mathcal{E}_{\theta}(H;\mathcal{D}) = 
\mathbb{E}_{x\sim\mathcal{D}}
\mathbb{E}_{\tau\sim\operatorname{Rollout}(H,\theta,x)}
\left[V(\tau,x)\right],
\end{equation}
and abbreviate $\mathcal{E}_{\theta}(H;\mathcal{D}^{\mathrm{test}})$ as
$\mathcal{E}_{\theta}(H)$. During search, the optimizer may request an
evaluation of $H$ on a subset
$Q\subseteq\mathcal{D}^{\mathrm{dev}}$ or
$Q\subseteq\mathcal{D}^{\mathrm{val}}$:
\begin{equation}
F_{\theta}(H,Q)\longrightarrow(\hat{s},\varphi),
\end{equation}
where $\hat{s}$ estimates aggregate performance on $Q$ (e.g. sample mean) and $\varphi$ contains per-case
outcomes and execution traces. A partition-specific disclosure policy
$\pi_{\mathcal{D}}$ determines which of these outputs the optimizer observes.
The test partition is inaccessible during search and is evaluated by the
evaluation server only after the optimizer nominates a candidate.

\paragraph{Budget.}
Each evaluation request $j$ incurs a non-negative cost vector
$c_j\in\mathbb{R}_{\geq 0}^{d}$, and search must satisfy $\sum_j c_j\leq B$
componentwise for a fixed budget vector $B$, so the optimizer chooses what to
evaluate and at what fidelity. \add{A separate budget vector $B_{\mathrm{opt}}$ can bound the optimizer's
own inference (for example, input tokens, output tokens, or model calls).}

\begin{figure}[t]
\centering
\includegraphics[width=\linewidth]{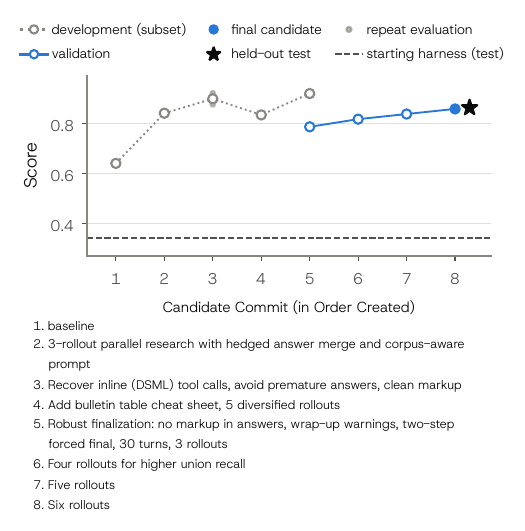}
\caption{\add{\textbf{One optimization trajectory} (\texttt{claude-opus-5} with \miniswe on
OfficeQA). Development and validation scores for each commit in order. Small dots are
repeat evaluations and larger points their mean. Commit 1 is the starting harness (held-out
score dashed). The filled point is the final candidate and the star its held-out score.}}
\label{fig:trajectory}
\end{figure}

\paragraph{Optimizer.}
An optimizer $O$ is any program satisfying
the interface of Algorithm~\ref{alg:protocol} that receives $H_0$, interacts with $F_{\theta}$
under disclosure $\pi$ and budget $B$, produces candidates
$H_1,\ldots,H_T\in\mathcal{H}$, and nominates a final candidate $H^{+}$.

\paragraph{Objective.}
The optimizer maximizes the expected improvement of its nominated candidate over
the pinned seed on the held-out partition,
\begin{align*}
\max_{H \in \mathcal{H}}\quad
\![
  \mathcal{E}_{\theta}(H) - \mathcal{E}_{\theta}(H_0)
]
\quad\text{s.t.}\quad
\sum_j c_j\leq \add{B .}
\end{align*}
Since $H_0$ is pinned, $\mathcal{E}_{\theta}(H_0)$ is a constant within a task. \add{Test data is never exposed, so} the optimizer never observes
the quantity it maximizes. Raw improvement may not be comparable across
tasks whose scoring scales differ, so wherever tasks are
compared we use normalized gain
\begin{equation}
g
=
\frac{
  \mathcal{E}_{\theta}(H^{+}) - \mathcal{E}_{\theta}(H_0)
}{
  1-\mathcal{E}_{\theta}(H_0)
},
\label{eq:normalized-gain}
\end{equation}
where negative values indicate a nominated candidate worse than the seed.

\subsection{The \heb suite}

\label{sec:suite}
\label{sec:scoring}

\add{Each \heb\ task pairs a benchmark with a pinned target model and a seed harness.}

\add{
\begin{table*}[t]
\centering\small
\setlength{\tabcolsep}{5pt}
\begin{tabular}{@{}ll r r r @{\hspace{11pt}}r @{\hspace{11pt}}r@{}}
\toprule
\textbf{Model} & \textbf{Coding Harness} & \textbf{OfficeQA} & \textbf{BrowseComp-Plus} & \textbf{Terminal-Bench} & \textbf{GAIA} & \textbf{Avg.\ Rank}\\
\midrule
claude-opus-5 & \texttt{claude-code} & $0.60$ {\tiny $\pm0.03$} & $0.46$ {\tiny $\pm0.05$} & $0.19$ {\tiny $\pm0.01$} & $0.44$ {\tiny $\pm0.05$} & $2.62$\\
claude-opus-5 & \texttt{mini-swe-agent} & $\mathbf{0.67}$ {\tiny $\pm0.06$} & $0.48$ {\tiny $\pm0.02$} & $0.19$ {\tiny $\pm0.04$} & $0.35$ {\tiny $\pm0.02$} & $2.88$\\
claude-opus-5 & \texttt{opencode} & $0.63$ {\tiny $\pm0.02$} & $\mathbf{0.55}$ {\tiny $\pm0.07$} & $\mathbf{0.28}$ {\tiny $\pm0.02$} & $0.41$ {\tiny $\pm0.07$} & $\mathbf{1.75}$\\
claude-sonnet-5 & \texttt{claude-code} & $0.57$ {\tiny $\pm0.04$} & $0.17$ {\tiny $\pm0.10$} & $0.08$ {\tiny $\pm0.05$} & $0.37$ {\tiny $\pm0.07$} & $7.88$\\
claude-sonnet-5 & \texttt{mini-swe-agent} & $0.54$ {\tiny $\pm0.02$} & $0.35$ {\tiny $\pm0.03$} & $0.11$ {\tiny $\pm0.05$} & $0.23$ {\tiny $\pm0.06$} & $9.00$\\
claude-sonnet-5 & \texttt{opencode} & $0.47$ {\tiny $\pm0.05$} & $0.24$ {\tiny $\pm0.09$} & $0.15$ {\tiny $\pm0.04$} & $0.26$ {\tiny $\pm0.01$} & $9.00$\\
gpt-5.6-sol & \texttt{codex} & $0.48$ {\tiny $\pm0.01$} & $0.11$ {\tiny $\pm0.07$} & $0.10$ {\tiny $\pm0.02$} & $\mathbf{0.45}$ {\tiny $\pm0.04$} & $8.50$\\
gpt-5.6-sol & \texttt{mini-swe-agent} & $0.49$ {\tiny $\pm0.04$} & $0.38$ {\tiny $\pm0.05$} & $0.09$ {\tiny $\pm0.03$} & $0.37$ {\tiny $\pm0.08$} & $7.12$\\
gpt-5.6-sol & \texttt{opencode} & $0.35$ {\tiny $\pm0.06$} & $0.17$ {\tiny $\pm0.10$} & $0.14$ {\tiny $\pm0.02$} & $0.32$ {\tiny $\pm0.02$} & $9.75$\\
gpt-5.6-terra & \texttt{codex} & $0.21$ {\tiny $\pm0.15$} & $0.09$ {\tiny $\pm0.11$} & $0.04$ {\tiny $\pm0.04$} & $0.25$ {\tiny $\pm0.06$} & $14.25$\\
gpt-5.6-terra & \texttt{mini-swe-agent} & $0.26$ {\tiny $\pm0.09$} & $0.37$ {\tiny $\pm0.02$} & $0.07$ {\tiny $\pm0.02$} & $0.15$ {\tiny $\pm0.01$} & $11.62$\\
gpt-5.6-terra & \texttt{opencode} & $0.24$ {\tiny $\pm0.12$} & $0.11$ {\tiny $\pm0.08$} & $0.06$ {\tiny $\pm0.03$} & $0.15$ {\tiny $\pm0.02$} & $13.88$\\
kimi-k3 & \texttt{kimi-cli} & $0.56$ {\tiny $\pm0.04$} & $0.33$ {\tiny $\pm0.09$} & $0.17$ {\tiny $\pm0.02$} & $0.29$ {\tiny $\pm0.02$} & $7.00$\\
kimi-k3 & \texttt{mini-swe-agent} & $0.56$ {\tiny $\pm0.04$} & $0.45$ {\tiny $\pm0.03$} & $0.18$ {\tiny $\pm0.03$} & $0.32$ {\tiny $\pm0.06$} & $5.00$\\
kimi-k3 & \texttt{opencode} & $0.47$ {\tiny $\pm0.12$} & $0.22$ {\tiny $\pm0.08$} & $0.14$ {\tiny $\pm0.05$} & $0.27$ {\tiny $\pm0.01$} & $9.75$\\
\bottomrule
\end{tabular}
\caption{Core-grid normalized gain (mean $\pm$ one standard error across $N{=}3$ independent optimization runs). The best result per task is bold. \oc and \miniswe are shared across models; the remaining condition is each model's native harness. Avg.\ Rank averages each configuration's rank among all 15 within each task (1 = best); it is descriptive and carries no resampled interval.}
\label{tab:gain-by-task}
\end{table*}
}

\paragraph{\add{Benchmarks.}}
\add{The suite uses four existing agent benchmarks: question answering
over enterprise documents in OfficeQA \citep{opsahlong2026officeqaproenterprisebenchmark},
deep research over a fixed corpus in BrowseComp-Plus \citep{browsecompplus2025},
command-line tasks in Terminal-Bench~2.0 \citep{merrill2026terminalbenchbenchmarkingagentshard},
and multi-step questions for general assistants in GAIA \citep{mialon2023gaia}.}
\add{Each benchmark's task environments and verifiers are used directly.} Environmental
constraints such as limited network access and strict wall clock bounds are used without
modification.

\paragraph{\add{Seed harnesses.}}
Every pinned seed $H_0$ across \heb tasks is a small, deliberately untuned Python
harness. Three of the four are competent but naive;
the OfficeQA seed, for example, is a $\sim$130-line agent built on the OpenAI API, with three
tools, a 24-turn loop, and a generic system prompt. GAIA's is a non-functional stub.

Table~\ref{tab:suite} compares the pinned \add{seeds} $H_0$ to \add{five off-the-shelf harnesses, existing agent harnesses run unmodified on each task,} showing that they perform comparably or worse than the weakest ones, providing ample headroom for improvement.

\paragraph{\add{Target models.}}
\add{During optimization,} $|\mathcal{M}| = 1$, i.e. we use one pinned target model per task; these are listed in Table~\ref{tab:suite}. Target models were chosen so that the seed would land in a measurable
mid-range rather than at the floor or the \add{ceiling. GAIA's stub is the exception and scores zero.}

\add{To test whether optimized harnesses generalize and optimizer performance depends on the target model, we evaluate final
candidates on two additional target models per task \emph{after optimization}, i.e. these transfer evaluations do not
affect the optimizer.}

\begin{figure}[t]
\centering
\includegraphics[width=\linewidth]{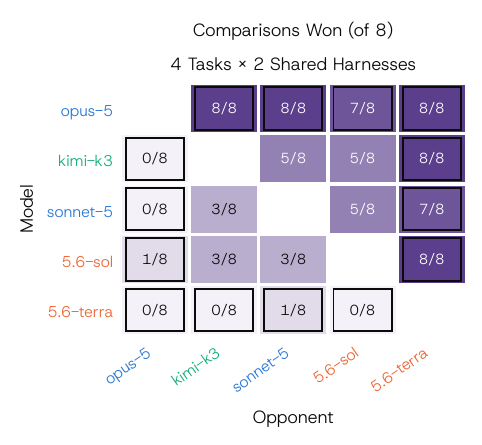}
\caption{\add{\textbf{Pairwise model comparison.} Each cell counts the comparisons, one per
task and shared harness (8 in all), in which the row model's mean gain ranks above the
column model's (ties count half). Outlined pairs are resolved (Section~}\ref{sec:analysis}\add{).}}
\label{fig:pairwise}
\end{figure}

\begin{figure*}[t]
\centering
\includegraphics[width=\linewidth]{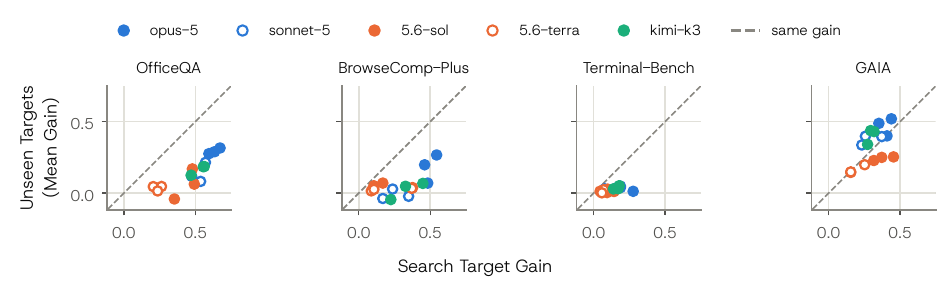}
\caption{\add{\textbf{Gain on unseen target models.} Each point is one optimizer
configuration's mean gain on the target model used during search (the search target) and
over the two unseen target models. Points below the dashed line keep less of their gain.}}
\label{fig:transfer-generalization}
\end{figure*}

\paragraph{\add{Scoring.}}
Each task's baseline $\mathcal{E}_{\theta}(H_0)$ is measured once, averaged over $K{=}3$ independent
rounds, and pinned for reproducibility. \add{The final} candidate is likewise
scored $K{=}3$ times per test case and averaged. \add{Each partition score averages over
every case in the partition. Cases whose runs time out or fail score zero.}

\paragraph{\add{Evaluation interface.}}
\add{In \heb, the optimizer can call the evaluation function up to 100 times on each of the
development and validation partitions and evaluate up to four times each partition's cases.
Total target model tokens are also capped.} Development reveals case inputs, per-case outcomes,
and traces to support diagnosis; validation reveals only an aggregate score to support
selection. 

\add{Every component of
$B_{\mathrm{opt}}$ is infinite, so optimizer inference is metered but not limited, and results
estimate what is achievable when the optimizer's compute is not the limiting factor.
Section~}\ref{sec:process}\add{ relates gain to optimizer compute.}

\paragraph{\add{Environment.}}
Figure~\ref{fig:arch} details our execution infrastructure. Each optimizer is \add{an LLM operating through a coding harness and} runs in an isolated sandbox with evaluation results, task data, and the target harness in the filesystem. Similarly, each target agent rollout is performed in an ephemeral sandbox to control for noise introduced by environmental drift.

\paragraph{\add{Example run.}}
\add{Figure~}\ref{fig:trajectory}\add{ shows one run. The optimizer first evaluates the
starting harness and its first candidate on a development subset, tests further candidates
on development cases, validates the most promising ones, and finally submits the candidate with the
best validation score. The submission is scored on the test partition only after search.
Appendix~}\ref{app:examples}\add{ shows two more example runs.}

\section{Experimental Setup}
\label{sec:setup}

\add{We run three sets of experiments. The \emph{core grid} crosses five optimizer models
with three coding harnesses each (two shared, one native) on all four tasks. The \emph{starting harness grids} and
\emph{target model transfer} evaluations test whether its conclusions hold with a different
starting harness or target model. The \emph{release series} runs earlier releases of two
model families to test whether gain tracks model progress.}

\begin{figure*}[t]
\centering
\includegraphics[width=\linewidth]{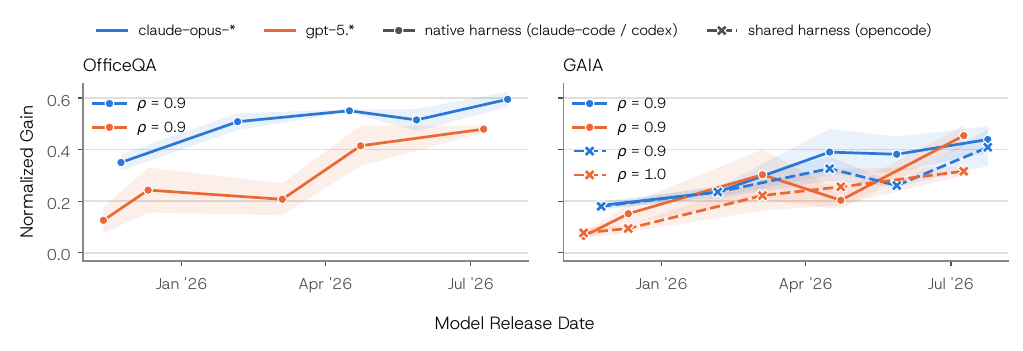}
\caption{\add{\textbf{Model release series.} Mean normalized gain over three runs per
release (shading: $\pm$ one standard error) by release date, from \texttt{claude-opus-4-5}
to \texttt{claude-opus-5} and from \texttt{gpt-5.1} to \texttt{gpt-5.6-sol}. Solid lines with
dots use each family's native harness (\cc, \codex); dashed lines with crosses use \oc. GAIA's
starting harness scores zero, so its normalized gain equals the held-out score. Legends give
each line's Spearman $\rho$ between release order and gain.}}
\label{fig:ladder}
\end{figure*}

\paragraph{\add{Core grid.}}
We evaluate five optimizer models from three developers on all four optimization tasks in our suite:
\texttt{claude-opus-5}, \texttt{claude-sonnet-5}, \texttt{gpt-5.6-sol},
\texttt{gpt-5.6-terra}, and \texttt{kimi-k3}, chosen \add{from} the current frontier
\add{releases of three developers}. Each model is paired with \add{three} coding harnesses: \add{two}
\textbf{shared \add{harnesses}} (\oc \add{and \miniswe}), held fixed across all models, and the model's
\textbf{native harness}, namely \cc for the Claude models, \codex for the GPT
models, and \kc for \add{\texttt{kimi-k3}}. We refer to each model--harness pair as an
\emph{optimizer configuration}\add{. Each configuration runs three times on each task, giving
\NumPrimaryConfigurations\ configurations, \NumPrimaryTaskCells\ task-specific cells, and
\NumPrimaryRuns\ runs} in \add{the core grid}. Comparing models under \add{a shared harness} holds the coding harness fixed; comparing a model
across \add{a shared harness} and its native harness measures sensitivity to \add{the coding harness}.

\paragraph{\add{Starting harness grids and target model transfer.}}
\add{On GAIA and Terminal-Bench, we run the full model--harness grid at three measured starting harness
strengths (\NumSeedGridRuns\ runs, including the core-grid starting harnesses). On GAIA, the starting harnesses are the core-grid
stub (0.000), the OpenCode agent (0.475), and a functional tool-using research agent (0.596).
On Terminal-Bench, they are a stub (0.000), the core-grid shell loop (0.241), and the OpenCode
agent (0.380). We also evaluate
core-grid final candidates on two additional target models per task (Section~}\ref{sec:suite}\add{).}

\paragraph{\add{Release series.}}
\add{For} two of the five models\add{, \texttt{claude-opus-5} and \texttt{gpt-5.6-sol},} we run earlier releases of the same family using each family's native harness on OfficeQA, \add{and on GAIA under both native and \oc harnesses
(\NumLadderAdditionalRuns\ runs beyond the core grid),}
\add{reported} in Section~\ref{sec:ladder}. \add{Each series spans five releases over eight months,
from November 2025 to July 2026.}

\paragraph{\add{Analysis.}}
\label{sec:analysis}
\add{We report the normalized gain $g$ of Eq.~}\ref{eq:normalized-gain}\add{, the fraction of the
headroom above the pinned baseline that an optimizer captures. Normalization is
task-specific and does not equalize measurement precision across tasks, so we compare
models by rank rather than by raw gain. Within each task and shared harness, we rank the
five models by mean gain, then average the ranks equally. Runs under native harnesses are
left out of this ranking: each family's native harness is different, so including them
would confound model with harness. Standard errors summarize variation across the three
runs of each configuration.}

\add{To decide whether one model ranks above another, we resample
the three runs of every configuration with replacement, recompute all ranks, and repeat
this \NumRankResamples\ times. We call a model pair \emph{resolved} when the 95\% interval of its
rank difference across these resamples excludes zero.}

\section{Results}
\label{sec:results}

Table~\ref{tab:gain-by-task} compares \add{the core grid} across tasks on normalized gain. Appendix~\ref{app:master} reports the same
results in \add{Tables~\ref{tab:master-officeqa}--\ref{tab:master-gaia-shell}}, together with \add{standard errors} and properties of
the generated harnesses by task.

\subsection{Distinguishing frontier models}
\label{sec:rq1}

The \add{model ranking resolves coarse differences but not} a complete
\add{ordering. Of \NumModelPairs\ model pairs, \NumResolvedModelPairs\ are resolved
(Figure~}\ref{fig:pairwise}\add{).
\texttt{claude-opus-5} ranks above each other model, and
\texttt{gpt-5.6-terra} ranks below each other model. The three middle models---}
\texttt{\add{kimi-k3}}, \texttt{claude-sonnet-5}, \add{and \texttt{gpt-5.6-sol}---remain
mutually unresolved.}

\add{Rank positions among the middle models also differ between \oc and \miniswe
(Appendix~}\ref{app:ranks}\add{), so close rankings remain conditional on the coding harness.}

\begin{figure*}[t]
\centering
\includegraphics[width=\linewidth]{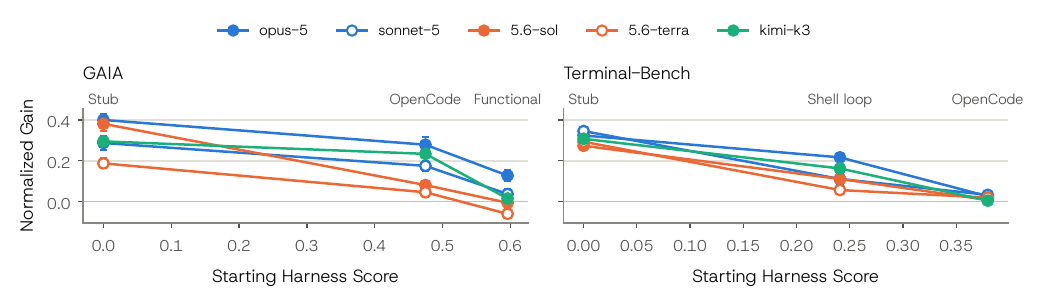}
\caption{\add{\textbf{Gain by starting harness.} Each model's normalized gain over its three
harness conditions and three runs (error bars: one standard error), at each starting harness's
measured score. OpenCode is used here as a starting harness, not as a coding harness.}}
\label{fig:seed-sensitivity}
\end{figure*}

\subsection{\add{Robustness to target} model}
\label{sec:transfer}

\leavevmode\add{We evaluated all \NumTransferPrimaryCandidates\ final candidates on two
additional target models per task. When gain is averaged over the target
model used during search and both additional targets, all \NumTransferRetainedPairs\ originally
resolved model pairs keep the same winner, no resolved pair reverses, and
\NumTransferResolvedPairs\ of \NumModelPairs\ pairs resolve in total
(Table~}\ref{tab:target-transfer}\add{).}

\add{Harness improvements carry over only partly (Figure~}\ref{fig:transfer-generalization}\add{).
On OfficeQA, BrowseComp-Plus, and Terminal-Bench, every final candidate keeps only part of
its gain on the unseen target models, although candidates with larger gains on the
original target tend to keep more. GAIA differs: its stub fails on every target, so any
working harness counts as gain (Appendix~}\ref{app:target-transfer}\add{).}

\subsection{Tracking model progress}
\label{sec:ladder}

\add{Within each release series, the task, budget, and coding harness are held fixed.}
\add{All \NumLadderSeries}\ \add{first-to-last comparisons resolve, and release-order
correlations range} from \add{\LadderRhoLo}\ to \add{\LadderRhoHi\
(Figure~}\ref{fig:ladder}\add{). At least} three of \add{four adjacent steps increase in
every series. The benchmark therefore tracks \add{progress across releases} in
these settings without implying monotonic improvement at every release. Over a model release timeline of
eight months, harness optimization improved along with general model development.}

\add{
\begin{table}[!htb]
\centering\small
\setlength{\tabcolsep}{5pt}
\resizebox{\columnwidth}{!}{%
\begin{tabular}{@{}llrrl@{}}
\toprule
\textbf{Benchmark} & \textbf{Starting harness} & \textbf{Score} & \textbf{Mean gain} & \textbf{Top model}\\
\midrule
GAIA & Stub & 0.000 & +0.310 & claude-opus-5\\
 & OpenCode & 0.475 & +0.163 & claude-opus-5\\
 & Functional & 0.596 & +0.023 & claude-opus-5\\
\addlinespace[3pt]
Terminal-Bench & Stub & 0.000 & +0.309 & claude-sonnet-5\\
 & Shell loop & 0.241 & +0.132 & claude-opus-5\\
 & OpenCode & 0.380 & +0.019 & claude-sonnet-5\\
\bottomrule
\end{tabular}
}
\caption{Summary by starting harness: its held-out score, mean normalized gain over the 45 runs, and the model with the highest mean gain. OpenCode is the OpenCode agent used as the starting harness, not as the optimizer's coding harness.}
\label{tab:seed-maturity}
\end{table}
}

\subsection{\add{Sensitivity to starting harness}}
\label{sec:seed-maturity}

\leavevmode\add{Starting harness strength changes both the available gain and some model-level
conclusions. Mean normalized gain decreases monotonically with starting harness score on both benchmarks
(Figure~}\ref{fig:seed-sensitivity}\add{; Table~}\ref{tab:seed-maturity}\add{), despite normalization for remaining headroom.
On Terminal-Bench, the top model also changes across starting harnesses. The ranking of
models therefore depends on the starting harness, not only the size of the gain.}

\subsection{Where current optimizers fall short}
\label{sec:process}

\add{Measures taken from the optimizers' trajectories describe} how \add{they} search; they are not
independent measures of held-out \add{performance.}

\begin{figure*}[t]
\centering
\includegraphics[width=\linewidth]{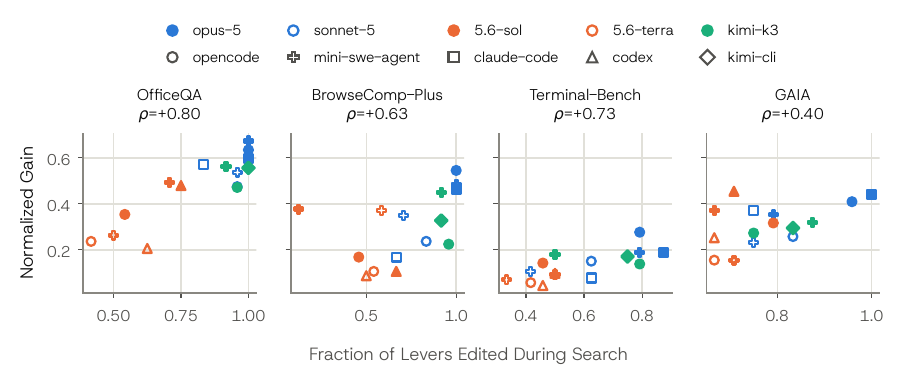}
\caption{\textbf{\add{Editing more levers} is associated with gain.} Normalized gain
against the fraction of eight pre-specified harness levers \add{edited} during
search. Each point averages \add{one configuration's runs; $\rho$ is computed per task.}}
\label{fig:levers}
\end{figure*}

\paragraph{Broader search is associated with greater gain.}
We identified eight harness levers on OfficeQA before examining the other tasks:
prompt, context management, step cap, retry and timeout policy, tool schema,
answer extraction, retrieval policy, and reasoning effort. The fraction \add{edited}
during search is positively associated with gain on every task, with
$\rho$ ranging from $\LeverRhoLo$ to $\LeverRhoHi$
(Figure~\ref{fig:levers}).

This \add{measures} \emph{exploration}\add{, not what is necessarily retained in the final candidate. The fraction of levers edited} is also
correlated with total modification volume, \add{which is itself positively associated
with gain on all four tasks ($\rho$ from \ChurnRhoLo\ to \ChurnRhoHi;
Appendix Figure~\ref{fig:edit-share-vs-gain}),} so the data
do not isolate \add{the number of levers edited} from search effort.

\paragraph{\add{Reading evaluation output} is not associated with higher gain.}
The share of actions spent reading evaluation output is negatively associated
with gain across the four tasks, from $\TraceRhoHi$ to $\TraceRhoLo$
\add{(Appendix Figure~\ref{fig:action-profile})}.
Optimizers rely mainly on per-case score summaries\add{, and full traces}
were requested only \NSpanCalls\ times by \NSpanCells\ of the
\add{\NumProcessRuns}\ \add{runs.} This does not establish that diagnosis is unnecessary.
For these tasks, per-case summaries may localize failures well enough that
reading full traces does not justify its context cost.

\paragraph{\add{Optimizer compute does not explain run-to-run variation.}}
\leavevmode\add{In a post hoc analysis, we compare the runs of each configuration on each
task, relating gain to the optimizer's output tokens on a log$_2$ scale
(Appendix~}\ref{app:compute}\add{). The common slope is
\ComputeGainPerOutputDoubling\ gain per token doubling, with a 95\% interval of
$[\ComputeGainLo,\ComputeGainHi]$ (Figure~}\ref{fig:opt-cost}\add{), so run-to-run
token variation within a fixed configuration has no detectable association
with gain. Optimizer compute was neither capped nor randomized, so this
does not rule out compute differences between models or harnesses.}

\paragraph{\add{Optimizers spend their evaluations in a few large batches.}}
\add{On the development and validation partitions, the optimizer can call the evaluation function
up to \EvalCallCapPerPartition\ times each (\EvalCallCapTotal\ in total) and evaluate up to four times
each partition's cases.} The median optimizer
\add{makes} \EvalCallsUsed\ calls (\EvalCallFrac\add{\ of the total cap)} but \add{uses}
\CasePassFrac\ of its case allowance, and \NBudgetExhausted\ of \add{\NBudgetRuns}\ \add{runs}
exhaust at least one partition's case \add{allowance. Optimizers run a few large evaluations
rather than many targeted ones.}

\paragraph{\add{Optimizers select on optimistic validation scores.}}
Most \add{runs} in Figure~\ref{fig:overfit} fall below the identity line: \add{for
\NumTestBelowValidation\ of the \NumValidationObserved\ runs with a validation
measurement,} the \add{final} candidate's test score is lower than the best validation score observed
during search. The figure cannot distinguish selection-induced
overfitting from a validation--test mismatch, so we claim only that the visible best score is
optimistic.

\subsection{The effect of the optimizer's own harness}
\label{sec:rq3}

Across the \add{\NumNativeOpenCodePairs}\ model--task pairs evaluated under both \add{native
and \oc harnesses, the native harness wins \NativeWinsOpenCode\ and} the shared
harness wins \add{\OpenCodeWins; against \miniswe}, the native harness wins
\add{\NativeWinsMiniSwe\ of \NumNativeMiniSwePairs\ pairs, loses \MiniSweWins}, and \add{ties
once (Appendix Figure~}\ref{fig:native-shared}\add{).} Native harnesses have no
consistent advantage. \add{Comparisons should therefore use shared harnesses rather than
treat a model's native tooling as part of the model.}

\section{Conclusion}
\heb makes \add{harness optimization} an empirical object: can a model diagnose,
modify, and improve an agent as measured via a held-out reward? Current
frontier models can, but unevenly. The strongest \add{separate coarsely from the rest,
and these distinctions persist across additional target models,} yet \add{close
rankings remain unresolved and conditional on the coding harness, and} gains
\add{depend strongly on the starting harness.} By releasing \heb, we make \add{end-to-end harness
optimization} reproducible to measure and concrete to optimize. The next frontier
is not merely better agents, but models that reliably make agents better.

\section*{Limitations}
\label{sec:limitations}
\heb is designed to be hack-resistant, not hackproof. The optimizer cannot access
the test partition or alter the target model, environment, or verifier, but repeated
development and validation feedback may still reward strategies specific to a
fixed evaluation. Future versions should introduce per-run jitter in cases, tool
behavior, and verifier implementation to distinguish general improvements from
exploitation of stable evaluator artifacts.

\add{\heb measures end-to-end improvement from given starting harnesses under a
specified feedback interface and budget. It does not decompose performance into
constituent skills, and it does not vary optimization strategy as a controlled
factor; the examples in Appendix~}\ref{app:examples} \add{illustrate strategies but do
not isolate their effects.}

The \add{starting} harness is itself a task-specific prior. \add{Three starting harnesses on GAIA}
and \add{Terminal-Bench show that gain falls as the starting harness matures,} but \add{they do} not
\add{span all starting harness architectures or domains.}

Candidates are restricted to Python and each task uses one pinned target
\add{model during search. Evaluating final candidates on two additional target models
tests post-search portability; it does not show that an optimizer would follow the
same trajectory if a different target model were used during search.}
Generalization to other languages, runtimes, \add{and} agent \add{architectures} remains
untested.

\add{Three runs per configuration and resampled rank intervals support
coarse pairwise distinctions, not a complete ordering of closely matched systems.
Optimizer inference was metered but not budget-matched: the within-configuration
output-token analysis detects no association with gain, but cannot remove
between-system compute confounding.}

\add{Cheaper proxy targets or predictive
development subsets may reduce evaluation cost, but their effect on optimizer
rankings remains to be tested.} Broader coverage, matched compute conditions, and
additional replication are needed before treating \heb as a comprehensive measure
of \add{harness optimization}.

\section*{Ethics Statement}

This work studies whether and how well LLMs can improve the code of other LLM
agents. Automating agent improvement carries a dual-use tension, since the same
capability that repairs and strengthens a benign agent could in principle be
turned toward a harmful one.

The benchmark's design limits this exposure. The
optimization target is always a bounded, benign downstream task (document QA,
deep research, multi-step reasoning, or terminal use), the score is the task's
own verifier on a held-out split and nothing else, and every optimizer runs inside
\add{a sandboxed environment} that meters spend, versions every change for audit, and
confines the target to a fixed model behind an allow-list -- a concrete
precaution given evidence that comparable agent benchmarks can be gamed via
evaluator hijacking or judge prompt-injection \citep{benchjack2026}. Nothing in
the setup rewards or requires acquiring new capabilities, tools, or resources
beyond editing a fixed agent's code.

Our benchmark reuses existing public datasets and benchmarks under their
respective licenses and adds \add{starting harnesses} and split definitions that we release.
The downstream corpora are public documents, and we introduce no personal or
sensitive data. \add{Agent evaluation is compute-intensive, so} we report token
usage as a first-class metric to make the cost of this line of work visible, and
we keep the \add{target-evaluation} budget denominated in evaluation calls and \add{case passes.} Finally, we report
the completed evaluation, including repeated runs, with explicit caveats (see
the Limitations section).

\bibliography{custom}

\clearpage

\appendix
\raggedbottom  
\etocdepthtag.toc{appendix}
\etocsettagdepth{main}{none}
\etocsettagdepth{appendix}{section}
\etocsettocstyle{\section*{\add{Appendix Contents}}}{}
\section{\add{Core results by task}}
\label{app:master}

\leavevmode\add{Tables~\ref{tab:master-officeqa}--\ref{tab:master-gaia-shell} give more detail on
the results in Table~\ref{tab:gain-by-task}, with one table per task. For each optimizer
configuration in the core grid, they report normalized gain and two further measures -- levers and target tokens --
each averaged over three independent optimization runs. Subscripts give standard
errors across those runs.}

\add{\emph{Levers} is the fraction of the eight pre-specified harness levers
(Section~\ref{sec:process}) that the optimizer edited at least once during search, whether
or not the edit was kept in the final candidate. \emph{Target tokens}
(TT) is the final candidate's target model token use per held-out
trial (i.e. one attempt at one test case), in millions. It counts every input token of every
model call, including tokens served from the provider's cache, as well as output tokens; input tokens make up nearly all of it.}

\add{Names are shortened to fit the tables: models drop their family prefix
(\texttt{opus-5} is \texttt{claude-opus-5}, \texttt{5.6-sol} is \texttt{gpt-5.6-sol}),
and \texttt{cc} and \texttt{mini-swe} denote \cc and \miniswe.}

\add{
\begin{table}[!htb]
\centering\footnotesize
\setlength{\tabcolsep}{2pt}
\begin{tabular}{@{}ll r r r@{}}
\toprule
\textbf{Model} & \textbf{Harness} & \textbf{Gain} & \textbf{Levers} & \textbf{TT (M)}\\
\midrule
opus-5 & \texttt{cc} & $0.60_{\pm0.03}$ & $1.00_{\pm0.00}$ & $2.57_{\pm1.53}$\\
opus-5 & \texttt{mini-swe} & $0.67_{\pm0.06}$ & $1.00_{\pm0.00}$ & $1.80_{\pm0.39}$\\
opus-5 & \texttt{opencode} & $0.63_{\pm0.02}$ & $1.00_{\pm0.00}$ & $1.09_{\pm0.19}$\\
sonnet-5 & \texttt{cc} & $0.57_{\pm0.04}$ & $0.83_{\pm0.11}$ & $1.65_{\pm0.26}$\\
sonnet-5 & \texttt{mini-swe} & $0.54_{\pm0.02}$ & $0.96_{\pm0.04}$ & $0.60_{\pm0.06}$\\
sonnet-5 & \texttt{opencode} & $0.47_{\pm0.05}$ & $0.96_{\pm0.04}$ & $0.77_{\pm0.09}$\\
5.6-sol & \texttt{codex} & $0.48_{\pm0.01}$ & $0.75_{\pm0.00}$ & $0.64_{\pm0.08}$\\
5.6-sol & \texttt{mini-swe} & $0.49_{\pm0.04}$ & $0.71_{\pm0.04}$ & $0.50_{\pm0.03}$\\
5.6-sol & \texttt{opencode} & $0.35_{\pm0.06}$ & $0.54_{\pm0.04}$ & $0.48_{\pm0.10}$\\
5.6-terra & \texttt{codex} & $0.21_{\pm0.15}$ & $0.62_{\pm0.07}$ & $0.57_{\pm0.09}$\\
5.6-terra & \texttt{mini-swe} & $0.26_{\pm0.09}$ & $0.50_{\pm0.07}$ & $0.46_{\pm0.07}$\\
5.6-terra & \texttt{opencode} & $0.24_{\pm0.12}$ & $0.42_{\pm0.04}$ & $0.49_{\pm0.08}$\\
kimi-k3 & \texttt{kimi-cli} & $0.56_{\pm0.04}$ & $1.00_{\pm0.00}$ & $0.83_{\pm0.12}$\\
kimi-k3 & \texttt{mini-swe} & $0.56_{\pm0.04}$ & $0.92_{\pm0.08}$ & $0.90_{\pm0.13}$\\
kimi-k3 & \texttt{opencode} & $0.47_{\pm0.12}$ & $0.96_{\pm0.04}$ & $1.00_{\pm0.15}$\\
\bottomrule
\end{tabular}
\caption{\textbf{OfficeQA} results by optimizer configuration (starting harness score $0.341$).}
\label{tab:master-officeqa}
\end{table}

\begin{table}[!htb]
\centering\footnotesize
\setlength{\tabcolsep}{2pt}
\begin{tabular}{@{}ll r r r@{}}
\toprule
\textbf{Model} & \textbf{Harness} & \textbf{Gain} & \textbf{Levers} & \textbf{TT (M)}\\
\midrule
opus-5 & \texttt{cc} & $0.46_{\pm0.05}$ & $1.00_{\pm0.00}$ & $3.08_{\pm0.92}$\\
opus-5 & \texttt{mini-swe} & $0.48_{\pm0.02}$ & $1.00_{\pm0.00}$ & $2.18_{\pm0.40}$\\
opus-5 & \texttt{opencode} & $0.55_{\pm0.07}$ & $1.00_{\pm0.00}$ & $4.00_{\pm0.31}$\\
sonnet-5 & \texttt{cc} & $0.17_{\pm0.10}$ & $0.67_{\pm0.15}$ & $4.31_{\pm0.91}$\\
sonnet-5 & \texttt{mini-swe} & $0.35_{\pm0.03}$ & $0.71_{\pm0.08}$ & $2.72_{\pm0.28}$\\
sonnet-5 & \texttt{opencode} & $0.24_{\pm0.09}$ & $0.83_{\pm0.04}$ & $4.51_{\pm1.00}$\\
5.6-sol & \texttt{codex} & $0.11_{\pm0.07}$ & $0.67_{\pm0.04}$ & $1.62_{\pm0.30}$\\
5.6-sol & \texttt{mini-swe} & $0.38_{\pm0.05}$ & $0.12_{\pm0.12}$ & $1.56_{\pm0.24}$\\
5.6-sol & \texttt{opencode} & $0.17_{\pm0.10}$ & $0.46_{\pm0.15}$ & $2.43_{\pm0.48}$\\
5.6-terra & \texttt{codex} & $0.09_{\pm0.11}$ & $0.50_{\pm0.00}$ & $2.42_{\pm0.17}$\\
5.6-terra & \texttt{mini-swe} & $0.37_{\pm0.02}$ & $0.58_{\pm0.08}$ & $2.33_{\pm0.17}$\\
5.6-terra & \texttt{opencode} & $0.11_{\pm0.08}$ & $0.54_{\pm0.04}$ & $2.72_{\pm0.10}$\\
kimi-k3 & \texttt{kimi-cli} & $0.33_{\pm0.09}$ & $0.92_{\pm0.08}$ & $4.10_{\pm1.29}$\\
kimi-k3 & \texttt{mini-swe} & $0.45_{\pm0.03}$ & $0.92_{\pm0.08}$ & $2.63_{\pm0.73}$\\
kimi-k3 & \texttt{opencode} & $0.22_{\pm0.08}$ & $0.96_{\pm0.04}$ & $4.59_{\pm0.56}$\\
\bottomrule
\end{tabular}
\caption{\textbf{BrowseComp-Plus} results by optimizer configuration (starting harness score $0.462$).}
\label{tab:master-browsecomp-plus}
\end{table}

\begin{table}[!htb]
\centering\footnotesize
\setlength{\tabcolsep}{2pt}
\begin{tabular}{@{}ll r r r@{}}
\toprule
\textbf{Model} & \textbf{Harness} & \textbf{Gain} & \textbf{Levers} & \textbf{TT (M)}\\
\midrule
opus-5 & \texttt{cc} & $0.19_{\pm0.01}$ & $0.88_{\pm0.00}$ & $1.46_{\pm0.28}$\\
opus-5 & \texttt{mini-swe} & $0.19_{\pm0.04}$ & $0.79_{\pm0.04}$ & $1.39_{\pm0.22}$\\
opus-5 & \texttt{opencode} & $0.28_{\pm0.02}$ & $0.79_{\pm0.04}$ & $1.30_{\pm0.22}$\\
sonnet-5 & \texttt{cc} & $0.08_{\pm0.05}$ & $0.62_{\pm0.07}$ & $0.83_{\pm0.36}$\\
sonnet-5 & \texttt{mini-swe} & $0.11_{\pm0.05}$ & $0.42_{\pm0.18}$ & $0.77_{\pm0.29}$\\
sonnet-5 & \texttt{opencode} & $0.15_{\pm0.04}$ & $0.62_{\pm0.07}$ & $1.56_{\pm0.41}$\\
5.6-sol & \texttt{codex} & $0.10_{\pm0.02}$ & $0.50_{\pm0.14}$ & $0.54_{\pm0.09}$\\
5.6-sol & \texttt{mini-swe} & $0.09_{\pm0.03}$ & $0.50_{\pm0.00}$ & $0.40_{\pm0.02}$\\
5.6-sol & \texttt{opencode} & $0.14_{\pm0.02}$ & $0.46_{\pm0.11}$ & $0.54_{\pm0.07}$\\
5.6-terra & \texttt{codex} & $0.04_{\pm0.04}$ & $0.46_{\pm0.08}$ & $0.43_{\pm0.06}$\\
5.6-terra & \texttt{mini-swe} & $0.07_{\pm0.02}$ & $0.33_{\pm0.11}$ & $0.74_{\pm0.04}$\\
5.6-terra & \texttt{opencode} & $0.06_{\pm0.03}$ & $0.42_{\pm0.04}$ & $0.56_{\pm0.12}$\\
kimi-k3 & \texttt{kimi-cli} & $0.17_{\pm0.02}$ & $0.75_{\pm0.00}$ & $0.85_{\pm0.18}$\\
kimi-k3 & \texttt{mini-swe} & $0.18_{\pm0.03}$ & $0.50_{\pm0.25}$ & $0.76_{\pm0.19}$\\
kimi-k3 & \texttt{opencode} & $0.14_{\pm0.05}$ & $0.79_{\pm0.04}$ & $0.99_{\pm0.35}$\\
\bottomrule
\end{tabular}
\caption{\textbf{Terminal-Bench} results by optimizer configuration (starting harness score $0.241$).}
\label{tab:master-terminal-bench}
\end{table}

\begin{table}[!htb]
\centering\footnotesize
\setlength{\tabcolsep}{2pt}
\begin{tabular}{@{}ll r r r@{}}
\toprule
\textbf{Model} & \textbf{Harness} & \textbf{Gain} & \textbf{Levers} & \textbf{TT (M)}\\
\midrule
opus-5 & \texttt{cc} & $0.44_{\pm0.05}$ & $1.00_{\pm0.00}$ & $0.83_{\pm0.25}$\\
opus-5 & \texttt{mini-swe} & $0.35_{\pm0.02}$ & $0.79_{\pm0.04}$ & $0.44_{\pm0.10}$\\
opus-5 & \texttt{opencode} & $0.41_{\pm0.07}$ & $0.96_{\pm0.04}$ & $0.45_{\pm0.24}$\\
sonnet-5 & \texttt{cc} & $0.37_{\pm0.07}$ & $0.75_{\pm0.00}$ & $0.24_{\pm0.07}$\\
sonnet-5 & \texttt{mini-swe} & $0.23_{\pm0.06}$ & $0.75_{\pm0.00}$ & $0.06_{\pm0.03}$\\
sonnet-5 & \texttt{opencode} & $0.26_{\pm0.01}$ & $0.83_{\pm0.04}$ & $0.11_{\pm0.03}$\\
5.6-sol & \texttt{codex} & $0.45_{\pm0.04}$ & $0.71_{\pm0.04}$ & $0.15_{\pm0.08}$\\
5.6-sol & \texttt{mini-swe} & $0.37_{\pm0.08}$ & $0.67_{\pm0.04}$ & $0.10_{\pm0.03}$\\
5.6-sol & \texttt{opencode} & $0.32_{\pm0.02}$ & $0.79_{\pm0.04}$ & $0.13_{\pm0.04}$\\
5.6-terra & \texttt{codex} & $0.25_{\pm0.06}$ & $0.67_{\pm0.04}$ & $0.10_{\pm0.06}$\\
5.6-terra & \texttt{mini-swe} & $0.15_{\pm0.01}$ & $0.71_{\pm0.04}$ & $0.03_{\pm0.01}$\\
5.6-terra & \texttt{opencode} & $0.15_{\pm0.02}$ & $0.67_{\pm0.04}$ & $0.04_{\pm0.01}$\\
kimi-k3 & \texttt{kimi-cli} & $0.29_{\pm0.02}$ & $0.83_{\pm0.04}$ & $0.12_{\pm0.00}$\\
kimi-k3 & \texttt{mini-swe} & $0.32_{\pm0.06}$ & $0.88_{\pm0.07}$ & $0.24_{\pm0.09}$\\
kimi-k3 & \texttt{opencode} & $0.27_{\pm0.01}$ & $0.75_{\pm0.07}$ & $0.18_{\pm0.02}$\\
\bottomrule
\end{tabular}
\caption{\textbf{GAIA} results by optimizer configuration (starting harness score $0.000$).}
\label{tab:master-gaia-shell}
\end{table}

}

\section{\add{Model rankings}}
\label{app:ranks}

\begin{figure}[h]
\centering
\includegraphics[width=\linewidth]{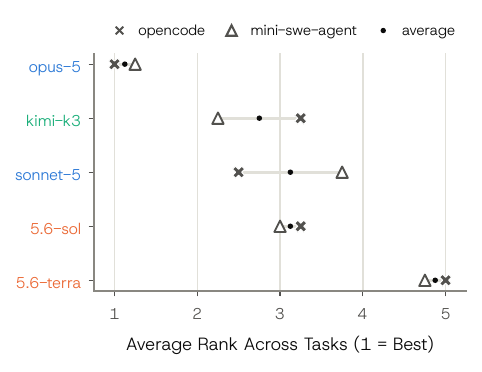}
\caption{\add{\textbf{Model ranks by shared harness.} Each model's rank among the five,
averaged over the four tasks, under \oc (crosses) and \miniswe (triangles);
dots mark the average of the two.}}
\label{fig:harness-ranks}
\end{figure}

Table~\ref{tab:model-effects} gives the \add{numeric ranks
behind Figures~\ref{fig:pairwise} and~\ref{fig:harness-ranks}. Models are ranked within each task and harness
before the ranks are averaged, so gains from tasks with different scales are never
compared directly.}

\add{
\begin{table}[!htb]
\centering\small
\setlength{\tabcolsep}{3pt}
\begin{tabular}{@{}lrrr@{}}
\toprule
\textbf{Model} & \textbf{\oc} & \textbf{\miniswe} & \textbf{Average}\\
\midrule
claude-opus-5 & 1.00 & 1.25 & 1.12\\
kimi-k3 & 3.25 & 2.25 & 2.75\\
claude-sonnet-5 & 2.50 & 3.75 & 3.12\\
gpt-5.6-sol & 3.25 & 3.00 & 3.12\\
gpt-5.6-terra & 5.00 & 4.75 & 4.88\\
\bottomrule
\end{tabular}
\caption{Model rankings. Each model is ranked among the five within each task under \oc and under \miniswe; ranks are averaged equally across tasks, and the last column averages the two harnesses. Lower is better. Seven of ten model pairs are resolved (Figure~\ref{fig:pair-intervals}).}
\label{tab:model-effects}
\end{table}
}

\add{Figure~}\ref{fig:harness-ranks}\add{ shows each model's rank under each shared harness. Rank
positions differ between \oc and \miniswe, particularly among the middle models, so a
single coding harness can overstate a fine-grained ordering.}

\begin{figure}[h]
\centering
\includegraphics[width=\linewidth]{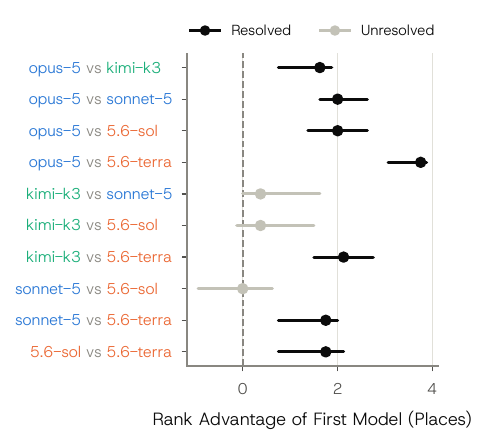}
\caption{\add{\textbf{Pairwise rank comparison.} For each model pair, how many places
higher the first-named model ranks on average over the eight task and shared-harness
combinations (dot), with a 95\% interval from resampling each configuration's three runs (line). Pairs
whose interval excludes zero are resolved (black).}}
\label{fig:pair-intervals}
\end{figure}

\add{Figure~\ref{fig:pair-intervals} shows, for every model pair, how many places higher
the better-ranked model ranks on average and the 95\% interval for that advantage.}

\section{\add{Optimizer sensitivity to the starting harness}}
\label{app:seed-maturity}

\leavevmode\add{The core grid starts each \add{benchmark} from one pinned starting harness. How much an optimizer
can add may depend on how capable that starting harness already is: a stub has to be
built into a working agent, while a mature agent leaves less headroom and is easier to
break. The starting harness grids therefore repeat the core-grid design on GAIA and
Terminal-Bench from three starting harnesses each, keeping everything else fixed.}

\add{Each starting harness grid uses the same five optimizer models, their three harness
conditions, and three independent runs per configuration: 45 runs per starting harness
and 270 in total, 90 of which are the core-grid runs. Benchmarks, splits, target model,
verifier, and budgets are those of the core grid; only the starting harness changes.
Gain is normalized against each starting harness's own measured score, so it is the
fraction of the remaining headroom an optimizer captures. Ordered by measured held-out
score, the starting harnesses are:}
\begin{itemize}[itemsep=1pt,topsep=2pt,leftmargin=*]
  \item \add{\textbf{GAIA.} \emph{Stub} (0.000): the core-grid starting harness, a minimal
  skeleton that does not answer questions. \emph{OpenCode} (0.475): the open-source
  OpenCode coding agent at a pinned release, used as the starting harness rather than as the
  optimizer's coding harness.
  \emph{Functional} (0.596): a working tool-using research agent that searches the web,
  inspects files and images, and runs shell commands.}
  \item \add{\textbf{Terminal-Bench.} \emph{Stub} (0.000): a minimal skeleton that does
  not solve tasks. \emph{Shell loop} (0.241): the core-grid starting harness, a function-calling shell
  loop with linear history and a fixed step budget. \emph{OpenCode} (0.380): the OpenCode
  agent built from source. It calls the target model through a generic Chat Completions
  client rather than OpenCode's built-in provider, runs more slowly than the stock \oc\ in
  Table~\ref{tab:suite}, and times out on 45 of 111 trials against 9, so it scores 0.380
  against 0.607.}
\end{itemize}

\add{Tables~\ref{tab:seed-grid-gaia} and~\ref{tab:seed-grid-terminal-bench} give every
configuration's gain from each starting harness, Figure~\ref{fig:seed-sensitivity} plots
each model's average, and Table~\ref{tab:seed-maturity} summarizes each grid. Gain falls
as the starting harness gets stronger on both benchmarks, but models differ in how fast it
falls. On GAIA, \texttt{claude-opus-5} keeps the most of its gain, while
\texttt{gpt-5.6-sol}, nearly level with it from the stub, loses most of it, and
\texttt{gpt-5.6-terra} on average makes the functional agent worse. On Terminal-Bench,
the models are close from the stub and from OpenCode and separate most from the shell
loop. The ordering of models therefore depends on the starting harness, not only the
size of the gain.}

\add{
\begin{table}[!htb]
\centering\footnotesize
\setlength{\tabcolsep}{3pt}
\begin{tabular}{@{}llrrr@{}}
\toprule
\textbf{Model} & \textbf{Harness} & \textbf{Stub} & \textbf{OpenCode} & \textbf{Functional}\\
\multicolumn{2}{@{}l}{\emph{starting harness score}} & \emph{0.000} & \emph{0.475} & \emph{0.596}\\
\midrule
opus-5 & \texttt{cc} & $0.44_{\pm0.05}$ & $0.38_{\pm0.01}$ & $0.19_{\pm0.02}$\\
opus-5 & \texttt{mini-swe} & $0.35_{\pm0.02}$ & $0.21_{\pm0.07}$ & $0.05_{\pm0.03}$\\
opus-5 & \texttt{opencode} & $0.41_{\pm0.07}$ & $0.26_{\pm0.06}$ & $0.14_{\pm0.05}$\\
sonnet-5 & \texttt{cc} & $0.37_{\pm0.07}$ & $0.18_{\pm0.04}$ & $0.04_{\pm0.03}$\\
sonnet-5 & \texttt{mini-swe} & $0.23_{\pm0.06}$ & $0.17_{\pm0.07}$ & $0.04_{\pm0.07}$\\
sonnet-5 & \texttt{opencode} & $0.26_{\pm0.01}$ & $0.18_{\pm0.02}$ & $0.04_{\pm0.01}$\\
5.6-sol & \texttt{codex} & $0.45_{\pm0.04}$ & $0.08_{\pm0.04}$ & $0.03_{\pm0.03}$\\
5.6-sol & \texttt{mini-swe} & $0.37_{\pm0.08}$ & $0.03_{\pm0.02}$ & $-0.07_{\pm0.03}$\\
5.6-sol & \texttt{opencode} & $0.32_{\pm0.02}$ & $0.13_{\pm0.03}$ & $0.03_{\pm0.05}$\\
5.6-terra & \texttt{codex} & $0.25_{\pm0.06}$ & $0.05_{\pm0.07}$ & $-0.02_{\pm0.04}$\\
5.6-terra & \texttt{mini-swe} & $0.15_{\pm0.01}$ & $0.05_{\pm0.03}$ & $-0.09_{\pm0.04}$\\
5.6-terra & \texttt{opencode} & $0.15_{\pm0.02}$ & $0.04_{\pm0.04}$ & $-0.07_{\pm0.01}$\\
kimi-k3 & \texttt{kimi-cli} & $0.29_{\pm0.02}$ & $0.26_{\pm0.01}$ & $0.06_{\pm0.01}$\\
kimi-k3 & \texttt{mini-swe} & $0.32_{\pm0.06}$ & $0.32_{\pm0.04}$ & $-0.03_{\pm0.03}$\\
kimi-k3 & \texttt{opencode} & $0.27_{\pm0.01}$ & $0.12_{\pm0.07}$ & $0.01_{\pm0.04}$\\
\bottomrule
\end{tabular}
\caption{\textbf{GAIA} normalized gain at each starting harness, by optimizer configuration (mean over three runs; subscripts are one standard error).}
\label{tab:seed-grid-gaia}
\end{table}

\begin{table}[!htb]
\centering\footnotesize
\setlength{\tabcolsep}{3pt}
\begin{tabular}{@{}llrrr@{}}
\toprule
\textbf{Model} & \textbf{Harness} & \textbf{Stub} & \textbf{Shell loop} & \textbf{OpenCode}\\
\multicolumn{2}{@{}l}{\emph{starting harness score}} & \emph{0.000} & \emph{0.241} & \emph{0.380}\\
\midrule
opus-5 & \texttt{cc} & $0.37_{\pm0.02}$ & $0.19_{\pm0.01}$ & $-0.00_{\pm0.03}$\\
opus-5 & \texttt{mini-swe} & $0.29_{\pm0.01}$ & $0.19_{\pm0.04}$ & $0.03_{\pm0.00}$\\
opus-5 & \texttt{opencode} & $0.31_{\pm0.02}$ & $0.28_{\pm0.02}$ & $0.05_{\pm0.02}$\\
sonnet-5 & \texttt{cc} & $0.35_{\pm0.02}$ & $0.08_{\pm0.05}$ & $0.02_{\pm0.03}$\\
sonnet-5 & \texttt{mini-swe} & $0.34_{\pm0.04}$ & $0.11_{\pm0.05}$ & $0.03_{\pm0.03}$\\
sonnet-5 & \texttt{opencode} & $0.35_{\pm0.01}$ & $0.15_{\pm0.04}$ & $0.04_{\pm0.02}$\\
5.6-sol & \texttt{codex} & $0.27_{\pm0.01}$ & $0.10_{\pm0.02}$ & $-0.01_{\pm0.02}$\\
5.6-sol & \texttt{mini-swe} & $0.28_{\pm0.01}$ & $0.09_{\pm0.03}$ & $0.01_{\pm0.01}$\\
5.6-sol & \texttt{opencode} & $0.27_{\pm0.01}$ & $0.14_{\pm0.02}$ & $0.02_{\pm0.00}$\\
5.6-terra & \texttt{codex} & $0.26_{\pm0.04}$ & $0.04_{\pm0.04}$ & $0.01_{\pm0.02}$\\
5.6-terra & \texttt{mini-swe} & $0.30_{\pm0.02}$ & $0.07_{\pm0.02}$ & $0.01_{\pm0.02}$\\
5.6-terra & \texttt{opencode} & $0.32_{\pm0.03}$ & $0.06_{\pm0.03}$ & $0.03_{\pm0.02}$\\
kimi-k3 & \texttt{kimi-cli} & $0.33_{\pm0.02}$ & $0.17_{\pm0.02}$ & $0.02_{\pm0.03}$\\
kimi-k3 & \texttt{mini-swe} & $0.37_{\pm0.02}$ & $0.18_{\pm0.03}$ & $0.01_{\pm0.03}$\\
kimi-k3 & \texttt{opencode} & $0.22_{\pm0.04}$ & $0.14_{\pm0.05}$ & $-0.01_{\pm0.03}$\\
\bottomrule
\end{tabular}
\caption{\textbf{Terminal-Bench} normalized gain at each starting harness, by optimizer configuration (mean over three runs; subscripts are one standard error).}
\label{tab:seed-grid-terminal-bench}
\end{table}

}

\section{\add{Harness generalization across target models}}
\label{app:target-transfer}

\leavevmode\add{Each task optimizes against one pinned target model, so a final candidate
may be tuned to that model's behaviour rather than to the benchmark. We test how far an
improvement carries over by scoring every core-grid final candidate, unchanged, on two
target models the optimizer never saw:}
\begin{itemize}[itemsep=1pt,topsep=2pt,leftmargin=*]
  \item \add{\textbf{OfficeQA and BrowseComp-Plus:} search target
  \texttt{deepseek-v4-flash}; unseen \texttt{gpt-5.4-mini} and \texttt{grok-build-0.1}.}
  \item \add{\textbf{Terminal-Bench:} search target \texttt{grok-build-0.1}; unseen
  \texttt{gpt-5.4-mini} and \texttt{gpt-oss-120b}.}
  \item \add{\textbf{GAIA:} search target \texttt{gpt-5.4-mini}; unseen
  \texttt{gpt-5.6-luna} and \texttt{grok-build-0.1}.}
\end{itemize}

\add{Each unseen-target evaluation uses the held-out test partition and the same scoring,
and its gain is normalized against the starting harness scored on that target; GAIA's stub scores
zero on every target. When a target rejects a harness's requests, for example because the
harness sends a parameter that target's API does not accept, the evaluation scores zero.
24 of the 356 unseen-target evaluations failed this way; the tables give the number of
such runs in parentheses. Two OfficeQA \texttt{gpt-5.6-terra} runs submitted the starting
harness unchanged; on any target that harness is the starting harness itself, so their unseen-target
gain is zero by definition and is counted as such.}

\add{Tables~\ref{tab:transfer-grid-officeqa}--\ref{tab:transfer-grid-gaia-shell} give
every configuration's gain on each target, Figure~\ref{fig:transfer-generalization}
compares gain on the search target with the average over the unseen targets, and
Table~\ref{tab:target-transfer} repeats the model ranking with all three targets.}

\add{On
OfficeQA, BrowseComp-Plus, and Terminal-Bench, every final candidate keeps only part of
its gain on unseen targets, although final candidates with larger search-target gains tend to
keep more, and those from \texttt{claude-opus-5} keep the most. GAIA differs because its stub fails
on every target, so any working harness counts as gain. On GAIA, almost all final candidates from the Claude and Kimi
models gain more on the unseen targets than on the search target, while every
\texttt{gpt-5.6-sol} final candidate and most \texttt{gpt-5.6-terra} final candidates are
rejected by one of the unseen targets. The coarse model ranking survives: averaging over all three targets keeps the
seven resolved pairs and resolves one more.}

\add{
\begin{table}[!htb]
\centering\footnotesize
\setlength{\tabcolsep}{1.7pt}
\begin{tabular}{@{}llrrr@{}}
\toprule
\textbf{Model} & \textbf{Harness} & \textbf{Search} & \multicolumn{2}{c}{\textbf{Unseen targets}}\\
 & & {\tiny\texttt{deepseek-v4-flash}} & {\tiny\texttt{gpt-5.4-mini}} & {\tiny\texttt{grok-build-0.1}}\\
\midrule
opus-5 & \texttt{cc} & $0.60_{\pm0.03}$ & $0.11_{\pm0.07}$ & $0.44_{\pm0.07}$\\
opus-5 & \texttt{mini-swe} & $0.67_{\pm0.06}$ & $0.17_{\pm0.13}$ & $0.46_{\pm0.09}$\\
opus-5 & \texttt{opencode} & $0.63_{\pm0.02}$ & $0.12_{\pm0.07}$ & $0.46_{\pm0.01}$\\
sonnet-5 & \texttt{cc} & $0.57_{\pm0.04}$ & $0.01_{\pm0.01}$ & $0.42_{\pm0.02}$\\
sonnet-5 & \texttt{mini-swe} & $0.54_{\pm0.02}$ & $0.03_{\pm0.00}$ & $0.13_{\pm0.02}$\\
sonnet-5 & \texttt{opencode} & $0.47_{\pm0.05}$ & $0.02_{\pm0.00}$ & $0.21_{\pm0.11}$\\
5.6-sol & \texttt{codex} & $0.48_{\pm0.01}$ & $0.06_{\pm0.03}$ & $0.27_{\pm0.08}$\\
5.6-sol & \texttt{mini-swe} & $0.49_{\pm0.04}$ & $0.04_{\pm0.01}$ & $0.09_{\pm0.07}$\\
5.6-sol & \texttt{opencode} & $0.35_{\pm0.06}$ & $-0.00_{\pm0.00}$ & $-0.08_{\pm0.06}$\\
5.6-terra & \texttt{codex} & $0.21_{\pm0.15}$ & $0.00_{\pm0.00}$ & $0.09_{\pm0.06}$\\
5.6-terra & \texttt{mini-swe} & $0.26_{\pm0.09}$ & $0.02_{\pm0.01}$ & $0.08_{\pm0.06}$\\
5.6-terra & \texttt{opencode} & $0.24_{\pm0.12}$ & $-0.01_{\pm0.01}$ & $0.04_{\pm0.09}$\\
kimi-k3 & \texttt{kimi-cli} & $0.56_{\pm0.04}$ & $0.04_{\pm0.01}$ & $0.33_{\pm0.08}$\\
kimi-k3 & \texttt{mini-swe} & $0.56_{\pm0.04}$ & $0.05_{\pm0.01}$ & $0.32_{\pm0.10}$\\
kimi-k3 & \texttt{opencode} & $0.47_{\pm0.12}$ & $-0.02_{\pm0.05}^{(1)}$ & $0.27_{\pm0.18}$\\
\bottomrule
\end{tabular}
\caption{\textbf{OfficeQA} gain on the search target and on two unseen target models (mean over runs; subscripts are one standard error).}
\label{tab:transfer-grid-officeqa}
\end{table}

\begin{table}[!htb]
\centering\footnotesize
\setlength{\tabcolsep}{1.7pt}
\begin{tabular}{@{}llrrr@{}}
\toprule
\textbf{Model} & \textbf{Harness} & \textbf{Search} & \multicolumn{2}{c}{\textbf{Unseen targets}}\\
 & & {\tiny\texttt{deepseek-v4-flash}} & {\tiny\texttt{gpt-5.4-mini}} & {\tiny\texttt{grok-build-0.1}}\\
\midrule
opus-5 & \texttt{cc} & $0.46_{\pm0.05}$ & $0.25_{\pm0.01}$ & $0.15_{\pm0.04}$\\
opus-5 & \texttt{mini-swe} & $0.48_{\pm0.02}$ & $0.07_{\pm0.05}$ & $0.07_{\pm0.07}$\\
opus-5 & \texttt{opencode} & $0.55_{\pm0.07}$ & $0.28_{\pm0.00}$ & $0.26_{\pm0.01}$\\
sonnet-5 & \texttt{cc} & $0.17_{\pm0.10}$ & $0.04_{\pm0.01}$ & $-0.11_{\pm0.03}$\\
sonnet-5 & \texttt{mini-swe} & $0.35_{\pm0.03}$ & $0.05_{\pm0.01}$ & $-0.09_{\pm0.01}$\\
sonnet-5 & \texttt{opencode} & $0.24_{\pm0.09}$ & $0.09_{\pm0.03}$ & $-0.03_{\pm0.04}$\\
5.6-sol & \texttt{codex} & $0.11_{\pm0.07}$ & $0.08_{\pm0.06}$ & $0.02_{\pm0.03}$\\
5.6-sol & \texttt{mini-swe} & $0.38_{\pm0.05}$ & $0.07_{\pm0.02}$ & $0.01_{\pm0.03}$\\
5.6-sol & \texttt{opencode} & $0.17_{\pm0.10}$ & $0.11_{\pm0.05}$ & $0.03_{\pm0.04}$\\
5.6-terra & \texttt{codex} & $0.09_{\pm0.11}$ & $0.05_{\pm0.03}$ & $-0.02_{\pm0.02}$\\
5.6-terra & \texttt{mini-swe} & $0.37_{\pm0.02}$ & $0.06_{\pm0.01}$ & $0.01_{\pm0.02}$\\
5.6-terra & \texttt{opencode} & $0.11_{\pm0.08}$ & $0.07_{\pm0.03}$ & $-0.02_{\pm0.03}$\\
kimi-k3 & \texttt{kimi-cli} & $0.33_{\pm0.09}$ & $0.12_{\pm0.03}$ & $-0.03_{\pm0.03}$\\
kimi-k3 & \texttt{mini-swe} & $0.45_{\pm0.03}$ & $0.12_{\pm0.02}$ & $0.02_{\pm0.04}$\\
kimi-k3 & \texttt{opencode} & $0.22_{\pm0.08}$ & $0.02_{\pm0.05}$ & $-0.11_{\pm0.03}$\\
\bottomrule
\end{tabular}
\caption{\textbf{BrowseComp-Plus} gain on the search target and on two unseen target models (mean over runs; subscripts are one standard error).}
\label{tab:transfer-grid-browsecomp-plus}
\end{table}

\begin{table}[!htb]
\centering\footnotesize
\setlength{\tabcolsep}{1.7pt}
\begin{tabular}{@{}llrrr@{}}
\toprule
\textbf{Model} & \textbf{Harness} & \textbf{Search} & \multicolumn{2}{c}{\textbf{Unseen targets}}\\
 & & {\tiny\texttt{grok-build-0.1}} & {\tiny\texttt{gpt-5.4-mini}} & {\tiny\texttt{gpt-oss-120b}}\\
\midrule
opus-5 & \texttt{cc} & $0.19_{\pm0.01}$ & $0.03_{\pm0.02}$ & $0.07_{\pm0.02}$\\
opus-5 & \texttt{mini-swe} & $0.19_{\pm0.04}$ & $0.03_{\pm0.00}$ & $0.05_{\pm0.02}$\\
opus-5 & \texttt{opencode} & $0.28_{\pm0.02}$ & $-0.01_{\pm0.02}$ & $0.03_{\pm0.02}$\\
sonnet-5 & \texttt{cc} & $0.08_{\pm0.05}$ & $0.03_{\pm0.01}$ & $0.03_{\pm0.01}$\\
sonnet-5 & \texttt{mini-swe} & $0.11_{\pm0.05}$ & $0.00_{\pm0.00}$ & $0.04_{\pm0.01}$\\
sonnet-5 & \texttt{opencode} & $0.15_{\pm0.04}$ & $0.03_{\pm0.02}$ & $0.03_{\pm0.02}$\\
5.6-sol & \texttt{codex} & $0.10_{\pm0.02}$ & $0.02_{\pm0.01}$ & $0.04_{\pm0.01}$\\
5.6-sol & \texttt{mini-swe} & $0.09_{\pm0.03}$ & $0.01_{\pm0.01}$ & $-0.00_{\pm0.02}$\\
5.6-sol & \texttt{opencode} & $0.14_{\pm0.02}$ & $0.00_{\pm0.03}$ & $0.02_{\pm0.00}$\\
5.6-terra & \texttt{codex} & $0.04_{\pm0.04}$ & $0.02_{\pm0.01}$ & $0.01_{\pm0.01}$\\
5.6-terra & \texttt{mini-swe} & $0.07_{\pm0.02}$ & $0.01_{\pm0.02}$ & $0.03_{\pm0.01}$\\
5.6-terra & \texttt{opencode} & $0.06_{\pm0.03}$ & $-0.00_{\pm0.02}$ & $0.01_{\pm0.01}$\\
kimi-k3 & \texttt{kimi-cli} & $0.17_{\pm0.02}$ & $-0.00_{\pm0.02}$ & $0.06_{\pm0.01}$\\
kimi-k3 & \texttt{mini-swe} & $0.18_{\pm0.03}$ & $0.06_{\pm0.01}$ & $0.04_{\pm0.00}$\\
kimi-k3 & \texttt{opencode} & $0.14_{\pm0.05}$ & $0.02_{\pm0.03}$ & $0.04_{\pm0.01}$\\
\bottomrule
\end{tabular}
\caption{\textbf{Terminal-Bench} gain on the search target and on two unseen target models (mean over runs; subscripts are one standard error).}
\label{tab:transfer-grid-terminal-bench}
\end{table}

\begin{table}[!htb]
\centering\footnotesize
\setlength{\tabcolsep}{1.7pt}
\begin{tabular}{@{}llrrr@{}}
\toprule
\textbf{Model} & \textbf{Harness} & \textbf{Search} & \multicolumn{2}{c}{\textbf{Unseen targets}}\\
 & & {\tiny\texttt{gpt-5.4-mini}} & {\tiny\texttt{gpt-5.6-luna}} & {\tiny\texttt{grok-build-0.1}}\\
\midrule
opus-5 & \texttt{cc} & $0.44_{\pm0.05}$ & $0.61_{\pm0.02}$ & $0.43_{\pm0.08}$\\
opus-5 & \texttt{mini-swe} & $0.35_{\pm0.02}$ & $0.57_{\pm0.03}$ & $0.41_{\pm0.03}$\\
opus-5 & \texttt{opencode} & $0.41_{\pm0.07}$ & $0.57_{\pm0.01}$ & $0.23_{\pm0.13}^{(1)}$\\
sonnet-5 & \texttt{cc} & $0.37_{\pm0.07}$ & $0.52_{\pm0.02}$ & $0.27_{\pm0.14}$\\
sonnet-5 & \texttt{mini-swe} & $0.23_{\pm0.06}$ & $0.43_{\pm0.07}$ & $0.24_{\pm0.14}^{(1)}$\\
sonnet-5 & \texttt{opencode} & $0.26_{\pm0.01}$ & $0.51_{\pm0.00}$ & $0.29_{\pm0.15}$\\
5.6-sol & \texttt{codex} & $0.45_{\pm0.04}$ & $0.51_{\pm0.03}$ & $0.00_{\pm0.00}^{(3)}$\\
5.6-sol & \texttt{mini-swe} & $0.37_{\pm0.08}$ & $0.50_{\pm0.08}$ & $0.00_{\pm0.00}^{(3)}$\\
5.6-sol & \texttt{opencode} & $0.32_{\pm0.02}$ & $0.46_{\pm0.04}$ & $0.00_{\pm0.00}^{(3)}$\\
5.6-terra & \texttt{codex} & $0.25_{\pm0.06}$ & $0.40_{\pm0.08}$ & $0.00_{\pm0.00}^{(3)}$\\
5.6-terra & \texttt{mini-swe} & $0.15_{\pm0.01}$ & $0.00_{\pm0.00}^{(3)}$ & $0.29_{\pm0.04}$\\
5.6-terra & \texttt{opencode} & $0.15_{\pm0.02}$ & $0.14_{\pm0.14}^{(2)}$ & $0.16_{\pm0.13}^{(1)}$\\
kimi-k3 & \texttt{kimi-cli} & $0.29_{\pm0.02}$ & $0.54_{\pm0.03}$ & $0.34_{\pm0.17}^{(1)}$\\
kimi-k3 & \texttt{mini-swe} & $0.32_{\pm0.06}$ & $0.54_{\pm0.04}$ & $0.32_{\pm0.16}^{(1)}$\\
kimi-k3 & \texttt{opencode} & $0.27_{\pm0.01}$ & $0.49_{\pm0.01}$ & $0.19_{\pm0.10}^{(1)}$\\
\bottomrule
\end{tabular}
\caption{\textbf{GAIA} gain on the search target and on two unseen target models (mean over runs; subscripts are one standard error).}
\label{tab:transfer-grid-gaia-shell}
\end{table}

}

\add{
\begin{table}[!htb]
\centering\small
\setlength{\tabcolsep}{5pt}
\begin{tabular}{@{}lrr@{}}
\toprule
\textbf{Model} & \textbf{Search target} & \textbf{All three targets}\\
\midrule
claude-opus-5 & 1.12 & 1.12\\
kimi-k3 & 2.75 & 2.50\\
claude-sonnet-5 & 3.12 & 3.00\\
gpt-5.6-sol & 3.12 & 3.75\\
gpt-5.6-terra & 4.88 & 4.62\\
\midrule
Resolved model pairs & 7/10 & 8/10\\
\bottomrule
\end{tabular}
\caption{Model rankings with gain on the search target alone and with gain averaged over the search target and the two unseen targets (lower is better). The last row counts resolved model pairs: all seven resolved on the search target keep the same winner, and one more resolves.}
\label{tab:target-transfer}
\end{table}
}

\section{\add{Sensitivity to optimizer compute}}
\label{app:compute}

\leavevmode\add{We ask whether runs that spend more optimizer compute reach higher gain. The
optimizer's own inference is metered but not capped (Section~\ref{sec:task}), so runs of
the same configuration used different amounts of it. For run $r$ of cell $c$
(one task and one optimizer configuration; 60 cells with three runs each), let
$x_{cr}$ be $\log_2$ of the optimizer's gateway-metered output tokens and $y_{cr}$ its
normalized gain. We subtract each cell's mean,
\[
  \tilde{x}_{cr} = x_{cr} - \bar{x}_c, \qquad \tilde{y}_{cr} = y_{cr} - \bar{y}_c,
\]
and fit one least-squares slope across all 180 runs,
\[
  \tilde{y}_{cr} = \beta\,\tilde{x}_{cr} + \varepsilon_{cr}.
\]

Centering removes everything that is constant within a cell, such as one model
being stronger or one task being easier, so $\beta$ compares only runs of the same
configuration on the same task. It is the change in gain for a run that used twice as many
output tokens as the other runs of its cell. The 95\% interval resamples cells with
replacement and refits the slope. Token use was not assigned by us, so $\beta$
describes an association; it does not show what would happen if an optimizer were given
more compute, and it does not equalize compute across models.}

\begin{figure}[!htb]
\centering
\includegraphics[width=\linewidth]{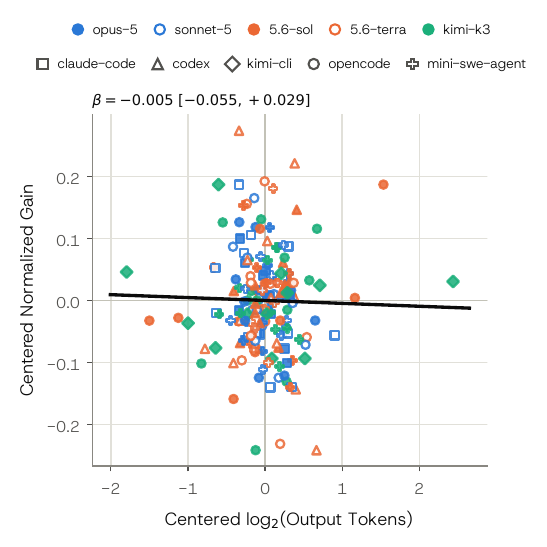}
\caption{\add{\textbf{Optimizer compute and gain.} Each point is one core-grid run,
centered on its configuration's mean on both axes; colour is the optimizer model and
shape its harness. The line is the fitted slope $\beta$,
given with its 95\% interval.}}
\label{fig:opt-cost}
\end{figure}

\section{\add{Illustrative optimization trajectories}}
\label{app:examples}

\begin{figure*}[t]
\centering
\includegraphics[width=\linewidth]{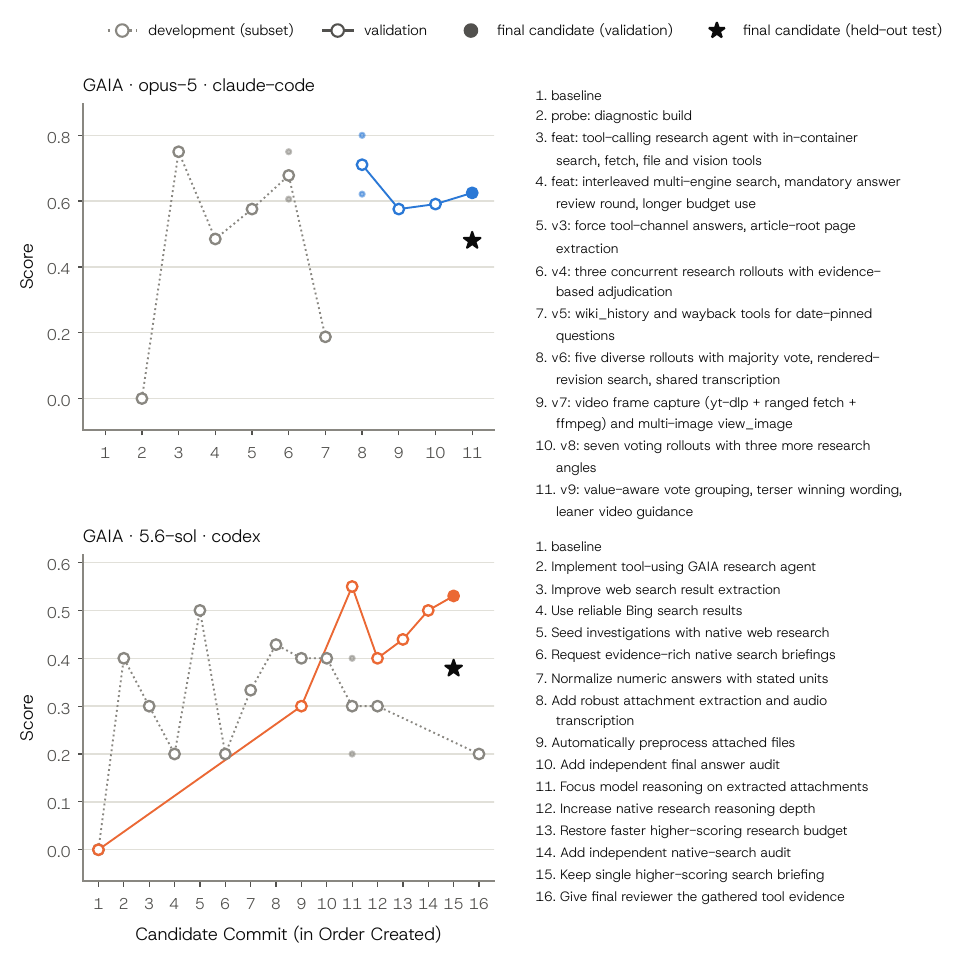}
\caption{\add{\textbf{Two optimization trajectories on GAIA.} Every candidate commit in the
order created, with its development (subset) and validation scores; the filled point is
the final candidate and the star its held-out test score. Commit messages are the
optimizer's own.}}
\label{fig:trajectories}
\end{figure*}

\leavevmode\add{Figure~\ref{fig:trajectories} follows two GAIA runs through every candidate the
optimizer committed, in the order it created them, with each commit message as the
optimizer wrote it. Development scores are measured on small case subsets and are noisy.
Validation scores use the full validation partition, and the star is the final candidate's
held-out test score. We chose these two runs because their commit messages describe the
changes most clearly, so they illustrate how optimizers search rather than a typical run.}

\add{Both runs start from GAIA's non-functional stub and first build a working tool-using
research agent. \texttt{claude-opus-5} then makes a few large redesigns: parallel research
rollouts with voting, archival search tools, and video handling. \texttt{gpt-5.6-sol}
makes many smaller changes to search and attachment handling, screening most of them on
development subsets before validating. In both runs the final candidate scores lower on
the held-out test than on validation.}

\add{\paragraph{A failure, the change, and the outcome.}
Figure~\ref{fig:example-case} follows one held-out BrowseComp-Plus test case from a
\texttt{claude-opus-5} run with \cc. The question asks for the title of a 1980s television
episode in which the hero helps the police arrest a smuggler with diplomatic immunity
(expected answer \emph{Illusion}, an episode of \emph{Manimal}). The starting harness's
search returns the first 512 words of each document. The title appears only deep inside a
fan blog whose opening is about another production. That blog is returned 25 to 44 times
per round, but the agent never opens it. In all three rounds it identifies the show and
the plot, reaches its 32-turn limit, and guesses a title. In its first commit the
optimizer changes search so that each hit also carries the passage that best matches the
query. On the same case, the first search of every test attempt returns the title passage,
and all three attempts answer correctly. The commit also adds multi-query search, a prompt
rewrite and a deadline guard, and no ablation separates their effects. The final harness
contains none of the benchmark's expected answers.}

\begin{figure*}[t]
\centering\footnotesize
\begin{minipage}[t]{0.49\linewidth}
\textbf{\add{Starting harness (fails in all three rounds)}}
\begin{verbatim}
turn 1   search: "1980s TV show episode main
         character helps police arrest smuggler
         diplomatic immunity"
  docid 31161  snippet: "--- title: Posts Tagged
    'Manimal' ... The timeless family classic,
    The Sound Of Music, continues to delight
    audiences ..."
  ...
turn 32  forced final answer:
  ... The episode title is "The Illusionist."
  Exact Answer: The Illusionist
verifier: incorrect (expected: Illusion)
\end{verbatim}
\textbf{\add{Final candidate (passes all three attempts)}}
\begin{verbatim}
turn 1   search: ["diplomatic immunity smuggler
         TV episode 1980s", ...]
  docid 31161
  lead:  "--- title: Posts Tagged 'Manimal' ...
          The Sound Of Music ..."
  match: "... Series: Manimal Episode: S1Ep2 -
          Illusion First Aired: 14 October 1983
          ... Jonathan is helping the police
          apprehend the man behind a smuggling
          ring ..."
  ...
  Exact Answer: Illusion
\end{verbatim}
\end{minipage}\hfill
\begin{minipage}[t]{0.49\linewidth}
\textbf{\add{Code change (excerpt of the new retriever.py)}}
\begin{verbatim}
def best_window(words, lowered, terms, size,
                skip):
    """Index of the size-word window covering
    the most distinct terms."""
    if not terms or len(words) <= skip + size:
        return skip, 0
    best_start, best_score = skip, -1
    step = max(10, size // 6)
    for start in range(skip, max(skip + 1,
                       len(words) - size + 1),
                       step):
        score = len(terms.intersection(
            lowered[start : start + size]))
        if score > best_score:
            best_start, best_score = start, score
    return best_start, best_score

def snippet_for(text, query, lead_words,
                window_words):
    words = text.split()
    lead = " ".join(words[:lead_words])
    ...
    start, score = best_window(words, lowered,
        terms, window_words, lead_words)
    ...
    return {"lead": lead, "match": "... "
        + " ".join(words[start : start
                         + window_words])
        + " ..."}
\end{verbatim}
\end{minipage}
\caption{\add{\textbf{One held-out test case before and after a harness change}
(BrowseComp-Plus case q1106, \texttt{claude-opus-5} with \cc). Left: the starting harness's
search shows only a document's opening and it never reads the title; the final candidate's
search adds the best-matching passage. Right: the retrieval change from the optimizer's
first commit, abridged. Traces are shortened and line-wrapped.}}
\label{fig:example-case}
\end{figure*}

\section{\add{Native harnesses give no consistent advantage}}
\label{app:harness-supplement}

\begin{figure*}[t]
\centering
\includegraphics[width=\linewidth]{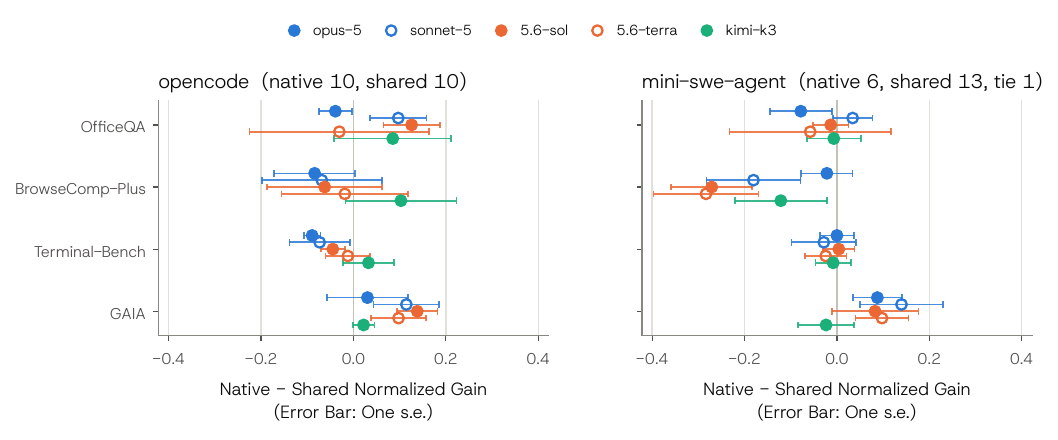}
\caption{\textbf{Native versus \add{shared coding harnesses.}} \add{Each point
is the native-minus-shared mean-gain difference for} one \add{task and model; error
bars show one standard error. Positive values favor the native harness.}}
\label{fig:native-shared}
\end{figure*}

\leavevmode\add{Each model runs in the core grid under two shared harnesses and under its own
native harness (\cc for the Claude models, \codex for the GPT models, \kc for
\texttt{kimi-k3}). If a model's native harness were simply better suited to it, the native
condition would win consistently. Figure~\ref{fig:native-shared} shows the native-minus-shared
difference in mean gain for every task and model. Native harnesses win
\NativeWinsOpenCode\ of \NumNativeOpenCodePairs\ comparisons against \oc and
\NativeWinsMiniSwe\ of \NumNativeMiniSwePairs\ against \miniswe. The direction depends
mostly on the task: native harnesses are ahead on GAIA against both shared harnesses for
most models, behind \miniswe on BrowseComp-Plus for all five models, and mixed on
OfficeQA and Terminal-Bench, where most differences are within one standard error.}

\section{\add{Search-process diagnostics}}
\label{app:process}

\leavevmode\add{These diagnostics support the search-process findings in
Section~\ref{sec:process}. They are observational: differences may reflect optimizer
model, harness, task, or search outcome rather than a causal mechanism.}

\subsection{\add{Optimizers mostly read harness code and rarely read evaluation output}}

\begin{figure*}[t]
\centering
\includegraphics[width=\linewidth]{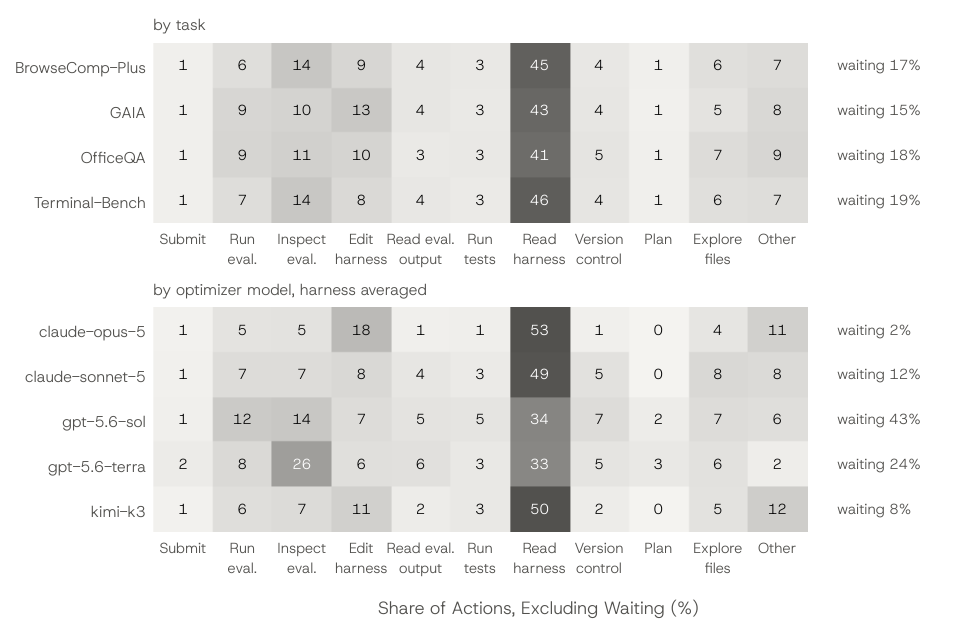}
\caption{\textbf{Optimizer action profiles.} \add{Share of actions in each category,
excluding waiting on evaluations, by task and by optimizer model (averaged over
harnesses); the share of all actions spent waiting is given at right.}}
\label{fig:action-profile}
\end{figure*}

\leavevmode\add{We classify every action an optimizer takes during search by what it does:
reading the harness code, editing it, running or inspecting an evaluation, reading
evaluation output (per-case results and traces), running tests, version control,
planning, or exploring files. Waiting on a running evaluation is counted separately,
because it inflates the action count without reflecting a decision.
Figure~\ref{fig:action-profile} gives the shares. Reading harness code is the largest
category on every task (41--46\% of actions), while reading evaluation output is
3--4\%, the share Section~\ref{sec:process} relates to gain. Models differ:
\texttt{claude-opus-5} spends 18\% of its actions editing, \texttt{gpt-5.6-terra} 26\%
inspecting evaluations, and \texttt{gpt-5.6-sol} spends 43\% of all its actions waiting
on evaluations.}

\subsection{\add{More modification goes with more gain}}

\begin{figure*}[t]
\centering
\includegraphics[width=\linewidth]{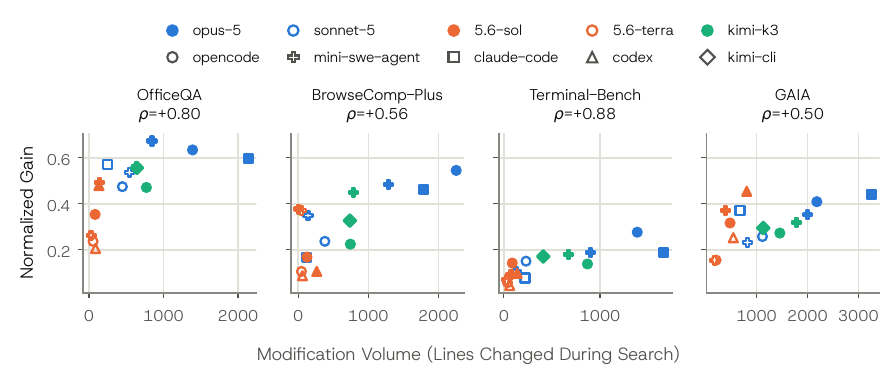}
\caption{\textbf{Modification volume and gain.} \add{Lines changed in the harness during
search against normalized gain; points average three runs per optimizer configuration,
and $\rho$ is computed within task.}}
\label{fig:edit-share-vs-gain}
\end{figure*}

\leavevmode\add{Figure~\ref{fig:edit-share-vs-gain} plots modification volume, the number of
lines the optimizer changed in the harness during search, against gain for each
optimizer configuration. The association is positive on all four tasks ($\rho$ from
\ChurnRhoLo\ to \ChurnRhoHi). The fraction of levers edited (Figure~\ref{fig:levers}) is correlated
with modification volume, so the two cannot be separated.}

\subsection{\add{On three tasks, higher-gain final candidates cost more to run}}

\begin{figure*}[t]
\centering
\includegraphics[width=\linewidth]{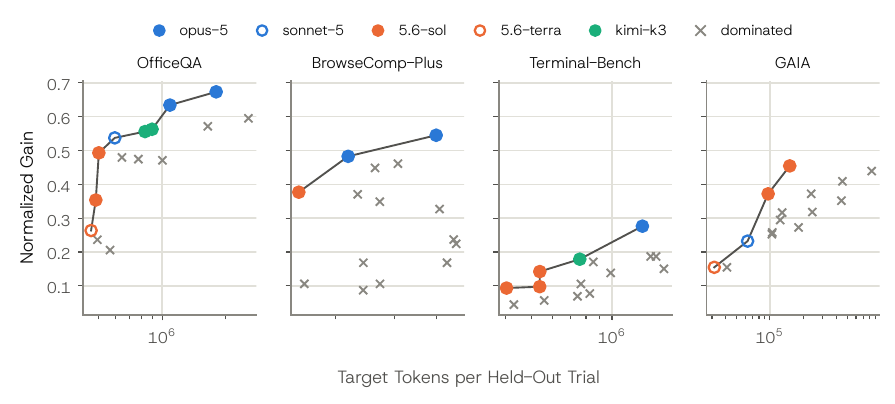}
\caption{\textbf{\add{Final candidate} cost and gain.} \add{Target tokens per held-out
trial against normalized gain for each optimizer configuration; filled points, joined
by a line, form the within-task cost--gain frontier, and crosses are configurations
beaten on both.}}
\label{fig:target-cost}
\end{figure*}

\leavevmode\add{Figure~\ref{fig:target-cost} plots each optimizer configuration's target tokens
per held-out trial (Appendix~\ref{app:master}) against its gain. Gain and target tokens
are positively associated on OfficeQA, Terminal-Bench, and GAIA ($\rho = 0.82$, $0.71$,
and $0.77$) but not on BrowseComp-Plus ($\rho = -0.03$). The line joins the
configurations that no other configuration beats on both gain and cost.}

\section{\add{Validation versus held-out scores}}
\label{app:selection}

\begin{figure}[!htb]
\centering
\includegraphics[width=\linewidth]{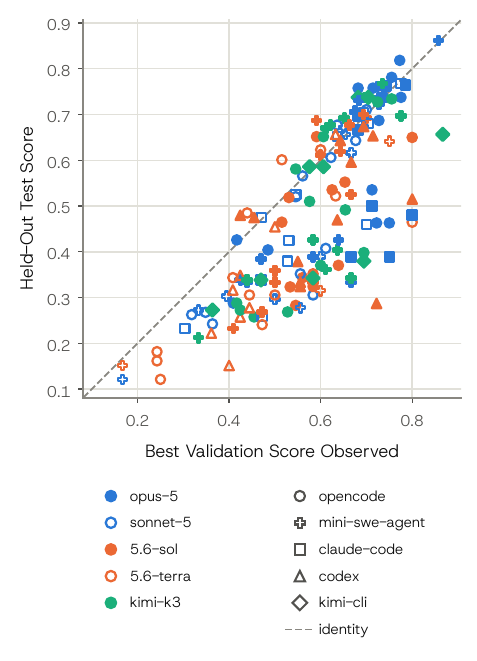}
\caption{\textbf{Validation versus held-out performance.} Best observed
validation score versus the final candidate's test \add{score for each core-grid
run;} the diagonal is equality. \add{Runs} without a validation measurement are omitted.}
\label{fig:overfit}
\end{figure}

\leavevmode\add{Validation measurements are available for 162 of 180 core-grid runs. In
Figure~}\ref{fig:overfit}\add{, 115 of those final candidates score below} the
best validation score observed during search.

\section{Reproducibility}
\label{app:repro}

Each task pins its dataset by immutable reference and its split by a committed
manifest. Splits are \add{non-overlapping,} and the split generator
verifies the committed tree byte-for-byte. Baselines are the starting harness's
held-out score pooled over $K{=}3$ rounds and are re-pinned whenever a starting harness
commit moves, using a script that reuses the original evaluation path.

Every
candidate an optimizer produces is an immutable Git commit, dependencies are
pinned by lockfile, and each evaluation runs in an isolated sandbox with
per-case timeouts taken from the dataset's own declared agent clock. The
gateway meters each evaluation's token usage as an input, cached, output, and
total split, and stamps request-log records so that per-trial token attribution
is reproducible from the run artifacts.

Every quantity we report about an
optimizer's process is recomputed from its own execution trace and from the
evaluator's record of which evaluations it ran, so the analysis is reproducible
from the released artifacts without re-running any search. The starting harnesses, split
manifests, build configurations, and
evaluation harness are released with the benchmark.

\end{document}